\documentclass[10pt,journal,compsoc]{IEEEtran}
\ifCLASSOPTIONcompsoc
  \usepackage[nocompress]{cite}
\else
  \usepackage{cite}
\fi
\ifCLASSINFOpdf
\else
\fi

\usepackage{booktabs}
\usepackage{amsfonts}
\usepackage{graphicx}
\usepackage{multirow}
\usepackage{algorithm}
\usepackage{algpseudocode}
\usepackage{amsmath}
\usepackage{caption}
\newtheorem{remark}{Remark}
\usepackage{color}
\usepackage{ragged2e}

\begin{document}
%
\title{HardBoost: Boosting Zero-Shot Learning with Hard Classes}
%
%
%
%

\author{Bo Liu, Lihua Hu, Zhanyi Hu, and Qiulei Dong
\IEEEcompsocitemizethanks{\IEEEcompsocthanksitem B. Liu and Z. Hu are with the National Laboratory of Pattern Recognition, Institute of Automation, Chinese Academy of Sciences, Beijing 100190, China, and also with the School of Future Technology, University of Chinese Academy of Sciences, Beijing 100049, China (e-mail: liubo2017@ia.ac.cn; huzy@nlpr.ia.ac.cn).
\IEEEcompsocthanksitem L. Hu is with the School of Computer Science and Technology, Taiyuan University of Science and Technology, Taiyuan 030024, China (e-mail: sxtyhlh@126.com).
\IEEEcompsocthanksitem Q. Dong is with the National Laboratory of Pattern Recognition, Institute of Automation, Chinese Academy of Sciences, Beijing 100190, China, also with the School of Artificial Intelligence, University of Chinese Academy of Sciences, Beijing 100049, China, and the Center for Excellence in Brain Science and Intelligence Technology, Chinese Academy of Sciences, Beijing 100190, China.(e-mail: qldong@nlpr.ia.ac.cn). Corresponding author: Qiulei Dong.}
}

%
%

\markboth{Submitted to IEEE TRANSACTIONS ON PATTERN RECOGNITION AND MACHINE INTELLIGENCE}%
{Shell \MakeLowercase{\textit{et al.}}: Bare Demo of IEEEtran.cls for Computer Society Journals}
%




\IEEEtitleabstractindextext{%
\begin{abstract}
\justifying This work is a systematical analysis on the so-called hard class problem in zero-shot learning (ZSL), that is, some unseen classes disproportionally affect the ZSL performances than others, as well as how to remedy the problem by detecting and exploiting hard classes. At first, we report our empirical finding that the hard class problem is a ubiquitous phenomenon and persists regardless of used specific methods in ZSL. Then, we find that high semantic affinity among unseen classes is a plausible underlying cause of hardness and design two metrics to detect hard classes. Finally, two frameworks are proposed to remedy the problem by detecting and exploiting hard classes, one under inductive setting, the other under transductive setting. The proposed frameworks could accommodate most existing ZSL methods to further significantly boost their performances with little efforts. Extensive experiments on three popular benchmarks demonstrate the benefits by identifying and exploiting the hard classes in ZSL. 
\end{abstract}

}

\maketitle

\IEEEdisplaynontitleabstractindextext

%
\IEEEpeerreviewmaketitle

\IEEEraisesectionheading{\section{Introduction}\label{sec:introduction}}

%
%
%
%
\IEEEPARstart{D}{eep} learning methods have achieved tremendous success in various fields, such as computer vision, natural language processing, and speech recognition. However, the success of deep learning methods heavily relies on a large number of labeled samples, and their power is constrained to handle those classes (seen classes) that have massive labeled samples for training. In fact, this demand is not easy to satisfy since labeling a great many samples is expensive, and new object classes are being created day by day. Hence, some learning methods which are able to deal with novel classes rapidly without costly training are expected eagerly, e.g. few-shot learning~\cite{wang2020generalizing}, zero-shot learning~\cite{Xian17Comprehensive} and so on. Among them, zero-shot learning (ZSL) which studies the most extreme case, i.e. to recognize novel classes (unseen classes) without any training labeled samples, has received much attention.

Since there are no labeled samples for unseen classes in the training, some auxiliary semantic information about unseen classes, e.g. expert-annotated attributes~\cite{Lampert14DAP}, pre-trained word embeddings~\cite{Frome13DeViSE}, or text description features~\cite{qiao2016ZSLNS}, are needed to make zero-shot learning (ZSL) \footnote{here we specialize in zero-shot visual classification} possible. The core idea of ZSL is to firstly learn visual-semantic knowledge from the seen-class domain by establishing an appropriate mapping between visual features and semantic features, then transfer the learned knowledge to the unseen-class domain. Usually, visual features are extracted by an existing convolutional neural network (CNN) backbone, e.g. ResNet~\cite{He16resnet}.

Many methods have been proposed to tackle the ZSL problem, which could be roughly divided into two groups: embedding based methods and generative methods. The embedding based ZSL methods~\cite{Xie_2019_AREN,liu2019LFGAA,dvbe2020,Li2020AJL} firstly mapped visual features and semantic features into a common embedding space, e.g. the visual feature space, or the semantic feature space, or an intermediate space, and then learned a mapping between the two kinds of features in the embedding space according to some similarity metrics. The generative ZSL methods~\cite{Xian18FCLSWGAN,schonfeld2019CADA-VAE,Narayan2020TF-VAEGAN,Long2018ZeroShotLU} tackled the ZSL problem by hallucinating unseen-class samples. They firstly employed a generative model to generate fake unseen-class samples conditioned on semantic features, then learned a classifier with these fake unseen-class samples to classify real unseen-class ones. In spite of the rapid development in the ZSL community, it is still an under-explored problem to design some high-efficiency training schemes to make use of the training samples for improving ZSL?

\begin{figure}[t]
	\centering
	\includegraphics[width=0.9\linewidth]{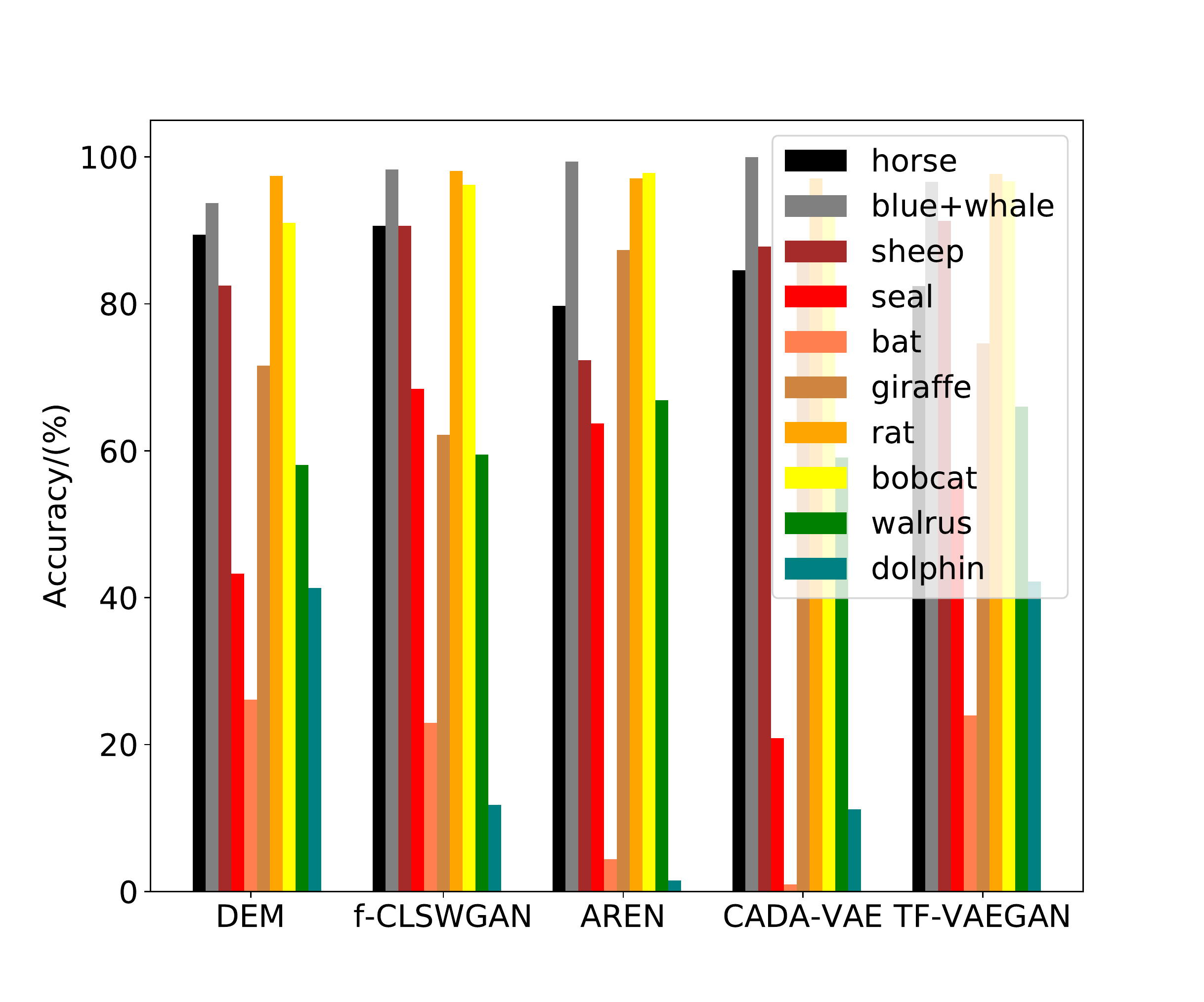}
	\caption{The hard class problem: for all the $5$ methods, some classes are consistently hard to classify.}
	\label{fig1}
\end{figure}

Addressing this problem, in this paper, we at first report an empirical finding that some unseen classes disproportionally affect the ZSL performances than others, seemingly independent of the used ZSL methods, coined `hard class problem' in this work, as shown in Figure~\ref{fig1}. Then, we conduct a systematical investigation on this problem from the following $3$ aspects: 1) What are the characteristics of hard classes in ZSL? 2) What is the possible cause of hard classes and how to identify them? 3) How to boost ZSL with hard classes? Targeting the first question, we empirically find that hard classes play a more important role than easy classes in ZSL, i.e., any knowledge on hard classes could bring notable benefits in ZSL. For the second aspect, based on confusion matrix analysis, we find that the misclassifications in ZSL are closely related to the similarities among classes in the semantic space, which indicates that high semantic affinity of some classes is a plausible cause of hardness among classes. Then, we propose two metrics to identify hard classes, the one is directly based on semantic similarity (SS) in the inductive setting, the other one is based on the class frequency (CF) of model predictions on unseen-class samples in the transductive setting. For the third aspect, we propose two frameworks to boost ZSL by exploiting hard-class information: 1) in the inductive setting, a \textbf{Har}dness based \textbf{S}ynthesizing framework (HarS) is proposed to synthesize `hard class' samples according to the proposed SS metric; 2) in the transductive setting, a \textbf{Har}dness based \textbf{S}electing \textbf{T}ransductive framework (HarST) is proposed to select pseudo-labeled unseen-class samples according to the CF metric for model self-training. Extensive experiments are conducted on three commonly used benchmarks to validate our proposed methods.

In sum, our contributions include:
\begin{itemize}
	\item We empirically find the existence of the hard class problem in ZSL, which is largely independent of the used ZSL methods.
	\item We find that the hardness of classes is closely related to their semantic similarities, and any knowledge on hard classes could benefit ZSL. Two metrics are proposed to identify hard classes.
	\item We propose two hardness based frameworks to boost ZSL in both the inductive and transductive settings. The proposed frameworks can accommodate most existing ZSL methods and improve their performances significantly with little efforts.
	\item Extensive experiments on three benchmarks demonstrate that the proposed two frameworks improve existing ZSL methods by large margins, and both could achieve the state-of-the-art performances.
\end{itemize}

The remaining part is organized as follows. Related works are reviewed in Section 2. The hard class problem is investigated from the above listed $3$ aspects in Section 3. Extensive experiments and discussions are reported in Section 4. Section 5 concludes the paper with some future directions.

Note that a preliminary version of this work was presented in~\cite{bo2021hardness}. Here the extensions include 1) generalizing the study of the hard class problem to the inductive ZSL; 2) providing an in-depth analysis on the possible cause of hard classes; 3) proposing a novel semantic similarity based metric for identifying hard classes in the inductive setting; 4) proposing a hardness based synthesizing framework to improve existing ZSL methods in the inductive setting; 5) conducting experiments on three benchmarks to validate the proposed methods.

\section{Related Work}
\subsection{Inductive ZSL}
In the inductive ZSL setting, only labeled seen-class data are available for model training. Existing works could be roughly categorized into two groups: embedding based methods and generative methods. 

The key of embedding based ZSL methods is to learn a discriminative embedding space where a mapping between visual features and semantic features could be established. According to the specific embedding space, the visual-to-semantic works~\cite{Akata16ALE,xian2016LATEM,socher2013CMT,Han_2021_CVPR,Fu2020VocabularyInformedZA} projected visual features into a semantic feature space and then related visual features with semantic ones in the semantic space according to some similarity metrics. By designing some novel attention modules~\cite{liu2021semantic,liu2019LFGAA}, semantically meaningful visual features could be learned from the input images to improve the visual-to-semantic mapping. However, such methods usually suffer from the hubness problem~\cite{zhang2017DEM}. The semantic-to-visual methods~\cite{Changpinyo17EXEM,zhang2017DEM,dvbe2020} could alleviate the hubness problem to some extent by mapping semantic features into a visual feature space via a linear~\cite{Changpinyo17EXEM} or non-linear~\cite{zhang2017DEM} regression model, which could be further improved with more advanced loss functions~\cite{dvbe2020,jiang2019TCN}. Some other works~\cite{Li2020AJL,Changpinyo16SYNC,Peng2018JointSA} projected both visual and semantic features into an intermediate embedding space, for instance, parameter space~\cite{Changpinyo16SYNC} or label space~\cite{Li2020AJL}, then learned the visual-semantic mapping in this space. 

The core idea of generative ZSL methods~ \cite{Xian18FCLSWGAN,NiZ019dascn,li2019LiGAN,paul2019SABR,zhu2019ABP,huang2019GDAN,schonfeld2019CADA-VAE,Zhu18GAZSL,verma2020ZSML,AFRNetLiuDH20,Shen2020IZF,LiuLY0L21,yue2021counterfactual} is to generate many fake unseen-class visual samples as similar to the real ones as possible, and then transform the ZSL problem into a conventional classification problem. This could be achieved by designing a conditional generative model whose conditions and outputs are semantic features and visual features respectively. The pioneering generative model based method~\cite{Xian18FCLSWGAN} employed a generative adversarial network (GAN) to generate the visual features, which was simultaneously regularized by a pre-trained feature classifier. Subsequently, many methods~\cite{NiZ019dascn,li2019LiGAN,paul2019SABR} were proposed to further regularize the conditional generator to generate better visual features, and other generative models like variational autoencoder (VAE)~\cite{schonfeld2019CADA-VAE} and normalizing flows (NF)~\cite{Shen2020IZF} were also employed to implement the feature generator. 

\subsection{Transductive ZSL}
In the transductive ZSL setting, besides the labeled seen-class data, the unlabeled unseen-class data are also used to train models. Kodirov et al.~\cite{Kodirov15uda} reshaped the transductive ZSL problem into an unsupervised domain adaptation (UDA) problem and proposed a UDA method with sparse coding to exploit the unlabeled unseen-class data. Fu et al.~\cite{Fu15TMV} proposed a hyper-graph based method, which firstly projected both visual features and semantic features into a multi-view embedding space and then constructed a hyper-graph to propagate labels for the unlabeled data. Song et al.~\cite{Song18QFSL} proposed an embedding learning based method. In this method, for the inputted unseen-class data, the sum of predictive probabilities on unseen classes was encouraged to be larger than those on seen classes. Besides, some iterative training based works~\cite{YeG19PREN, li2019GXE,bo2021hardness,liu2021iterative} were also proposed to take advantage of these unlabeled unseen-class data, where models were learned iteratively by firstly pseudo-labeling unseen-class samples and then re-training the model together with the pseudo-labeled data. Recently, several GAN based methods~\cite{Narayan2020TF-VAEGAN,paul2019SABR,xian2019f-VAEGAN,wu2020SDGN} were proposed. In these methods, a shared conditional generator was used to generate seen-class and unseen-class samples and two discriminators were respectively used to discriminate fake/real seen-class and unseen-class samples.

\section{Methodology}
In this section, we introduce some necessary definitions about zero-shot learning (ZSL) at first. Then, our empirical finding of the hard class problem in ZSL is described, followed by our systematical investigation on the hard class problem in both the inductive and transductive settings, organized by progressively answering the three connected questions: 1) What are the characteristics of hard classes? 2) What is the plausible cause of hard classes and how to identify them? 3) How to boost ZSL by exploiting hard classes?

\subsection{Definitions}
Generally speaking, ZSL is to learn a model to classify unseen-class data when only seen-class data are labeled for model training. Formally, at the training stage, we are given a labeled seen-class dataset $\mathcal{D}^{S}_{tr} = \{(x_{n}, y_{n})\}_{n=1}^{N}$, where $x_{n}$ is the $n$-th visual sample and $y_{n}$ is the corresponding class label, and $N$ is the number of samples in $\mathcal{D}^{S}_{tr}$. Usually, $x_{n}$ is a feature vector in $\mathbf{R}^{v}$ and $y_{n}$ is a scalar which belongs to the seen-class label set $Y^{S}$. In addition, a semantic feature set $\mathcal{E}=\{e_{y} \in \mathbf{R}^{s} \mid y \in Y\}$ which represents the semantic information of every class in $Y$ is also given, where $Y$ is the total class label set which includes not only the seen-class label set $Y^{S}$ but also the unseen-class label set $Y^{U}$. Note that $Y^{S}$ is disjoint with $Y^{U}$. At the testing stage, given a test set $X$, in the ZSL setting, the task is to learn a mapping $F_{ZSL}: X \to Y^{U}$ with the training set $\mathcal{D}^{S}_{tr}$ and the semantic feature set $\mathcal{E}$, while in the generalized ZSL (GZSL) setting, the task is to learn a mapping $F_{GZSL}: X \to Y$.  We denote the unlabeled unseen-class dataset and unlabeled seen-class dataset at the testing stage by $\mathcal{D}^{U}$ and $\mathcal{D}^{S}$ respectively. In the transductive ZSL setting, only the testing unseen-class dataset $\mathcal{D}^{U}$ is available for training. While in the transductive GZSL setting, both the testing unseen-class dataset $\mathcal{D}^{U}$ and the testing seen-class dataset $\mathcal{D}^{S}$ are available for training.

\subsection{The Hard Class Problem}
In this paper, we investigate $5$ representative ZSL methods: DEM~\cite{zhang2017DEM}, AREN~\cite{Xie_2019_AREN}, f-CLSWGAN~\cite{Xian18FCLSWGAN}, CADA-VAE~\cite{schonfeld2019CADA-VAE}, and TF-VAEGAN~\cite{Narayan2020TF-VAEGAN}, where DEM and AREN are two embedding based ZSL methods while f-CLSWGAN, CADA-VAE and TF-VAEGAN are three generative ZSL methods. Specifically, we compute the per-class accuracies of the unseen classes in the public AWA2~\cite{Xian17Comprehensive} dataset with the $5$ ZSL methods. The results are shown in Figure~\ref{fig1}. From Figure~\ref{fig1}, we observed a novel phenomenon: some unseen classes have very high accuracies while some others have quite low ones. Importantly, this phenomenon seems independent of the used methods. In other words, given some unseen classes in a ZSL task, they could be generally categorized into two groups: the hard group and the easy group. The difficulty of ZSL mainly lies in the handling of the hard group. In this work, those unseen classes with relatively low accuracies are named \textbf{hard classes} while those with relatively high accuracies are named \textbf{easy classes}. Note that similar problems also exist on other datasets, we do not present them here due to the limited space. Since this phenomenon has not been reported in the literature to our best knowledge, we term this problem as the \textbf{hard class problem}. In the following, we will systematically investigate this problem.

\begin{figure*}[t]
	\centering
	\includegraphics[width=0.9\linewidth]{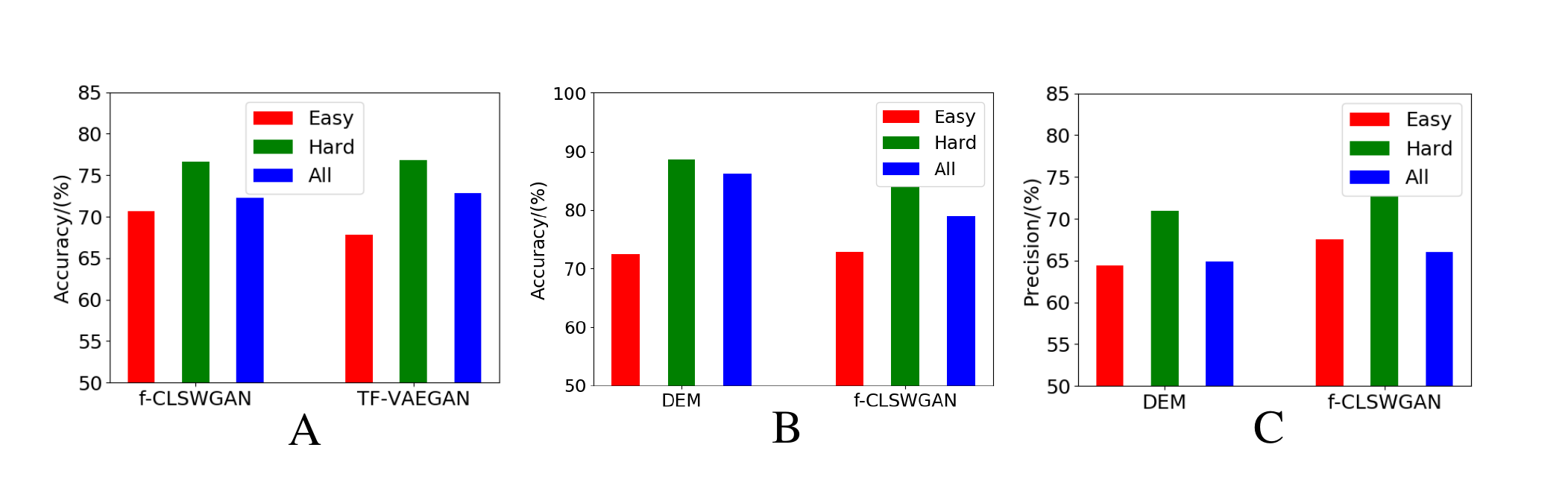}
	\caption{A: The ACCs of the models respectively trained with more easy-class samples, more hard-class samples, and all-unseen-class samples in the inductive setting. B: The ACCs of the models respectively re-trained with additional easy-class, hard-class and all-unseen-class data in the transductive setting. C: The precisions of the pseudo labels of easy classes, hard classes, and all unseen classes.}
	\label{fig2}
\end{figure*}

\subsection{What Are the Characteristics of Hard Classes?}
The first question interesting us is what are the characteristics of hard (and easy) classes. To answer this question, here we firstly make an assumption that the true hard (or easy) unseen classes are known as an oracle. Then, we investigate the effect of training models with hard-class (or easy-class) information on ZSL performances. By such contrastive analysis, we could reveal the characteristics of hard (and easy) classes. The experiments and analysis are conducted in both the inductive and transductive ZSL settings in the following.

\subsubsection{Inductive ZSL Setting}
Here the goal is to analyze the characteristics of hard (and easy) classes in the inductive ZSL training process by conducting experiments on AWA2 with f-CLSWGAN~\cite{Xian18FCLSWGAN} and TF-VAEGAN~\cite{Narayan2020TF-VAEGAN}. f-CLSWGAN and TF-VAEGAN are two generative ZSL methods, i.e. they firstly generate many unseen-class visual samples and then learn a classifier with these generated samples to classify real unseen-class ones. Here we employ f-CLSWGAN and TF-VAEGAN for analysis because it is straightforward to inject hard-class information into the generative ZSL methods by varying the number of generated unseen-class samples of hard (or easy) classes to train the classifier, and the effects can also be easily observed by evaluating the classifier's performances. Specifically, for both f-CLSWGAN and TF-VAEGAN, we firstly split the $10$ unseen classes in AWA2 into $5$ easy classes and $5$ hard ones by computing and ranking their per-class accuracies, i.e. the $5$ classes with the Top-$5$ highest accuracies are regarded as easy classes, the remaining $5$ as hard classes. Then, we generate three groups of fake unseen-class data: 1) the one including $N_{1}$ samples per hard class and $2*N_{1}$ per easy class; 2) the one including $N_{1}$ samples per easy class and $2*N_{1}$ per hard class; 3) the one including $1.5*N_{1}$ samples per class. Note that the three groups of data have the same number of samples so that the influence of training sample number could be removed in this analysis. Finally, the three groups of data are used to train three classifiers respectively and their performances are evaluated by average per-class accuracy (ACC). We denote the three classifiers by `easy', `hard', and `all' accordingly. The results are reported in Figure~\ref{fig2}-A. As shown in Figure~\ref{fig2}-A, for both f-CLSWGAN and TF-VAEGAN, the `hard' classifier achieves the best performance and it exceeds the `easy' one by a large margin. These results indicate that hard classes are more beneficial than easy classes to learn the inductive ZSL models.

\subsubsection{Transductive ZSL Setting}
Here we investigate the characteristics of hard (and easy) classes in the transductive ZSL setting. To this end, we conduct experiments on AWA2 with two representative ZSL methods, DEM (an embedding based ZSL method) and f-CLSWGAN (a generative ZSL method). In transductive ZSL, considering that unlabeled unseen-class samples are available for training, we could investigate the characteristics of hard (or easy) classes by firstly adding the hard-unseen-class (or easy-unseen-class) samples to the training dataset to re-train the models respectively, and then evaluating their performances. However, this scheme faces with the problem that true labels of individual unseen-class samples are actually unavailable for model re-training. What are available are only their pseudo labels predicted by a ZSL model trained with the labeled seen-class data. Addressing this problem, the following two sub-problems are investigated: 1) we firstly assume that true labels of individual unseen-class samples are available, so that the effect of hard (or easy) classes could be investigated; 2) then we analyze the quality of the pseudo labels of hard (and easy) classes, and observe the gap between pseudo labels and true labels. If the pseudo labels of hard (or easy) classes are accurate enough, then they are expected to replace the true ones as a sub-optimal alternative.

To investigate the first sub-problem, for both DEM and f-CLSWGAN, we firstly split the $10$ unseen classes in AWA2 into $5$ easy ones and $5$ hard ones as done in the inductive setting. Next, we construct three groups of unseen-class data by: 1) randomly selecting $N_{2}$ samples from each easy class; 2) randomly selecting $N_{2}$ samples from each hard class; 3) randomly selecting $N_{2}/2$ samples from each unseen class. By such, the three groups of data have the same number of samples. Note that these selected unseen-class data are attached with their true labels (assumed to be available). Then, the three groups of unseen-class data are added into the labeled seen-class dataset to re-train three ZSL models respectively. For clarity, we denote the three ZSL models by `easy', `hard', and `all' accordingly. Finally, their performances are evaluated by ACC. The results are shown in Figure~\ref{fig2}-B, where we can see that the `hard' model outperforms both `easy' and `all' significantly for both DEM and f-CLSWGAN, which demonstrates that the hard classes play a more important role than easy classes in the transductive training of ZSL models.

The above observation is achieved under the assumption that true labels of individual unseen-class samples are available. However, what are really available are only their pseudo labels. Here we investigate the quality of the pseudo labels of hard (or easy) classes. Specifically, we firstly obtain the $5$ easy classes and the $5$ hard ones in AWA2 as done before. Meanwhile, their pseudo labels are obtained by the ZSL model trained with the labeled seen-class dataset. Then, we evaluate the quality of their pseudo labels by computing their average per-class precision (PRE, i.e. the ratio of the true positives to all the predicted positives). The results are shown in Figure~\ref{fig2}-C. As shown in Figure~\ref{fig2}-C, the precision of the pseudo labels of hard classes is significantly higher than those of easy classes and all unseen classes, mainly because the number of the predicted pseudo labels for a hard class is much smaller than the number of the samples belonging to this class. This demonstrates that the pseudo-label quality of hard classes is better than that of easy classes. 

Based on the results in the above two experiments, we conclude that using hard-class samples in the transductive ZSL training is better than using easy-class ones owning to not only their larger impact but also their cleaner pseudo labels.

With the above experiments in both the inductive and transductive settings, we empirically demonstrate that hard unseen classes play a more important role than easy ones in ZSL. Moreover, it indicates that proper exploitation of hard-class information has a huge potential benefit to boost ZSL performance even with some straightforward ways and marginal efforts. However, to tap this potential, we need to find some ways to identify hard classes. We will explore this issue in the next section.

\begin{figure*}[t]
	\centering
	\includegraphics[width=0.9\linewidth]{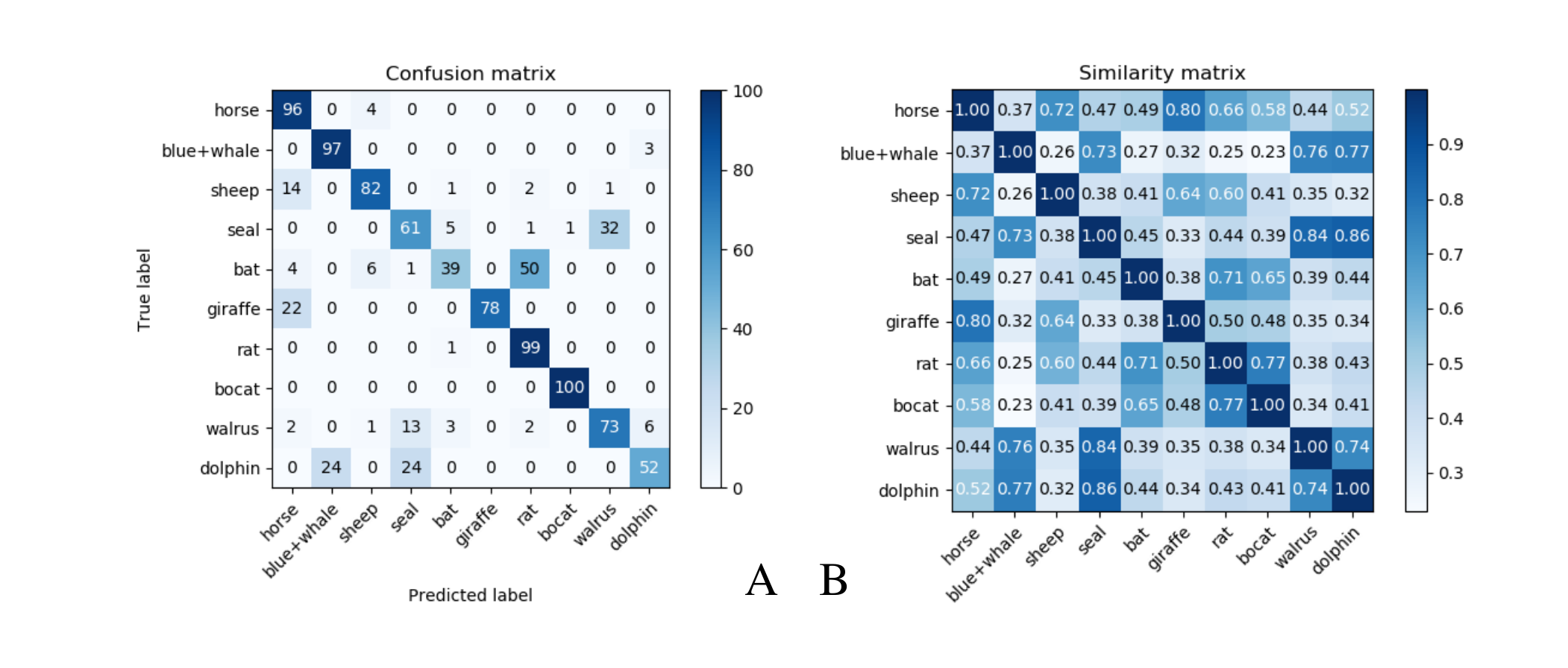}
	\caption{A: Confusion matrix of unseen classes in AWA2 with DEM. B: Semantic similarity matrix of unseen class in AWA2.}
	\label{fig3}
\end{figure*}

\subsection{How to Identify Hard Classes}
In this section, we will firstly analyze possible causes of the hard class problem from the perspective of semantic affinity, then design two metrics to identify hard classes.

\subsubsection{Possible Cause of Hard Classes}
Here, we firstly compute the confusion matrices of unseen classes in AWA2 by $5$ representative ZSL methods: DEM~\cite{zhang2017DEM}, AREN~\cite{Xie_2019_AREN}, f-CLSWGAN~\cite{Xian18FCLSWGAN}, CADA-VAE~\cite{schonfeld2019CADA-VAE}, and TFVAEGAN~\cite{Narayan2020TF-VAEGAN}. Note that $100$ samples per class are sampled to compute confusion matrices for reducing the influence of data unbalance. Since the confusion matrices of the $5$ methods are generally similar, here we only show the confusion matrix of DEM in Figure~\ref{fig3}-A and the others can be found in the supplementary materials. 

As seen from Figure~\ref{fig3}-A, many hard-class samples are usually misclassified into an unseen class which is semantically similar to its real class. For instance, `seal' samples are mainly misclassified into `walrus', and vice versa. Intuitively, this indicates that the cause of hard classes is probably related to the semantic affinities between classes. Based on this observation, we then compute the cosine similarity matrix of unseen classes in the semantic feature space, which is shown in Figure~\ref{fig3}-B. To quantitatively describe the correlation between confusion matrix and the semantic similarity matrix, we design two measurements from different perspectives. The one is the average per-class recall (APR), which is computed as follows: for each unseen class $c$, according to the confusion matrix, the $k$ classes into which the class $c$ is misclassified with the Top-$k$ largest frequencies, are regard as the true confusing classes of the class $c$. According to the similarity matrix, the $k$ classes which have the Top-$k$ largest semantic similarities with the class $c$, are taken as the predicted confusing classes of the class $c$. Then, the recall of the class $c$'s confusing classes is computed, and finally APR is obtained.
	
The other one is the average per-class misclassification rate (AMR) computed as follows: for each class $c$, according to the similarity matrix, we compute the Top-$k$ most similar classes of the class $c$. Then in the confusion matrix, we count the number ($n_{s}$) of samples which actually belong to the class $c$ but are misclassified into its most similar $k$ classes, and also count the number $n_{m}$ of the class $c$'s misclassifications. Finally, the misclassification rate of the class $c$ is denoted by $\frac{n_{s}}{n_{m}}$, and AMR is obtained by averaging these per-class ones.

The results of APR on AWA2 are $0.67$ and $0.78$ when $k=1$ and $k=2$ respectively, and AMRs are $0.51$ and $0.88$ when $k=1$ and $k=2$ respectively. These results suggest that semantic affinities among unseen classes are predictive of their misclassifications. In other words, semantic similarity of unseen classes is a useful indicator of hard classes. 

\subsubsection{Inductive ZSL Setting---SS Metric}
Here we study how to use semantic similarity to identify hard classes in the inductive ZSL setting. We propose a \textbf{S}emantic \textbf{S}imilarity based metric, called the SS metric. Its core idea is that for an unseen class $c$, if: 1) its semantic feature is considerably similar with that of another unseen class; 2) its semantic feature is considerably dissimilar with those of all seen classes, then this unseen class $c$ is most likely to be a hard class. The first condition indicates that the class $c$ is easy to be confused with another unseen class, while the second one implies that the learned ZSL model with the seen-class data knows little about the unseen class $c$, hence it would be hard for the model to recognize this class. 

Accordingly, for a given unseen class $c$, the proposed SS metric is computed by the following three steps: 1) computing the cosine similarity distances (1-cosine similarity) between the class $c$ and other unseen classes, recording the smallest one as $d_{c}^{u}$; 2) computing the cosine similarity distances between the class $c$ and all seen classes, recording the mean of the Top-$3$ smallest ones as $d_{c}^{s}$; 3) computing the SS metric $d_{c}$ by:
\begin{equation}
	d_{c} = d_{c}^{u}-d_{c}^{s} \\
	\label{eq1}
\end{equation}
The computation of the SS metric $d_{c}$ indicates that a smaller $d_{c}$ means a harder class. Hence, we finally identify the $K$ hard classes by choosing the Top-$K$ ranking classes from the following set:
\begin{equation}
	\label{eq2}
	\{h_{1}, \cdots, h_{i}, \cdots, h_{C}\} = \mathrm{argsort} (\{d_{1}, \cdots, d_{c}, \cdots, d_{C}\})
\end{equation}
where $h_{i}$ is the $i$-th hardest unseen class, $C$ is the number of unseen classes, and $\mathrm{argsort}(\cdot)$ ranks the inputs in the ascending order and returns the indexes of sorted inputs.

\subsubsection{Transductive ZSL Setting---CF Metric}
Here our goal is to study how to identify hard classes in the transductive ZSL setting. In Section 3.4.2, since only the semantics of unseen classes are available in the inductive setting, a semantics-based metric for identifying hard classes is proposed. However, in the transductive setting, (unlabeled) unseen-class visual samples are also available for training. Considering that the visual features of unseen classes usually contain richer information than their semantic features, here we propose a new hard class identification metric for the trasnductive setting, which is based on the class frequency of model predictions on these unseen-class visual features. The general idea of this metric is that when we predict pseudo labels for unseen-class visual features, if an unseen class $c$ is rarely predicted by the model, this implies that the model has seen little data from semantically similar classes with the class $c$, then the class $c$ is very likely to be a hard one. 

Specifically, an iterative self-training framework is employed in the transductive setting. At each iteration, assume that a set of pseudo labels are predicted on unseen-class data by the model, denoted as $P$. Then, we compute a histogram about these pseudo labels $P$ by
\begin{equation}
	\label{eq3}
	\{f_{1}, \cdots, f_{c}, \cdots, f_{C}\} = \mathcal{H}(P)
\end{equation}
where $f_{c}$ is the frequency of the pseudo-label class $c$, $C$ is the number of unseen classes, and $\mathcal{H}(\cdot)$ is the histogram function. Then, we regard the $K$ classes with the Top-$K$ lowest frequencies as hard classes:
\begin{equation}
	\label{eq4}
	\{h_{1}, \cdots, h_{i}, \cdots, h_{C}\} = \mathrm{argsort} (\{f_{1}, \cdots, f_{c}, \cdots, f_{C}\})
\end{equation}
where $h_{i}$ is the $i$-th hardest unseen class, and $\mathrm{argsort}(\cdot)$ is the ascending ranking function which returns the indexes. Since this metric is based on the class frequency, it is named the CF metric. 

Here we would like to give some more words on the CF metric. Although the design of CF does not directly rely on semantic similarity, it is implicitly connected to the semantic similarity. This is because deep neural networks are usually prone to predict the inputs as the classes that they have seen frequently in the training, if an unseen class is semantically similar to many of the training classes, then this class would be easy to classify by this model to some extent, and at the same time, the model would predict many inputs as this class since this class is similar to many training classes. On the contrary, if an unseen class is rarely predicted by the model, it is very likely that the model has seen few training classes which are semantically similar to this class, consequently would be hard for the model. Some supporting results on this reasoning are provided in the supplementary materials, which shows that the hard classes identified by the CF metric partly overlap with those identified by the SS metric. 
\begin{remark}
	Note that we also investigated other three approaches to identify hard classes in the transductive setting for comparison: 1) using the hard classes identified by the SS metric; 2) using the intersection set of hard classes identified by the SS and CF metrics; 3) using the union set of hard classes identified by the SS and CF metrics. We find that the CF metric is the most suitable one in the transductive setting. Due to the limited space, the results and analysis about the other 3 approaches are provided in the supplementary materials. 
\end{remark}

However, the CF metric has a shortcoming since it only considers the model predictions but not the datasets. Suppose we have an extremely class-unbalanced dataset, where an easy unseen class $c$ has very few samples. Then, the class frequency $f_{c}$ of the class $c$ computed only based on model predictions would be relatively small. i.e., the class $c$ would be wrongly identified as a hard class according the CF metric. To alleviate this problem, we propose an improved version of CF, named prior normalized class frequency (PnCF) metric. For the PnCF metric, the class frequency $f_{c}$ of the class $c$ is firstly normalized by the corresponding class prior $p_{c}$, as $\hat{f}_{c} = \frac{f_{c}}{p_{c}}$, and then we replace the $f_{c}$ in Equation~\ref{eq4} with $\hat{f}_{c}$ to identify hard classes. However, class priors are usually unknown for a given dataset. Hence, we further propose a classification-clustering consistency based method (3C) to estimate class priors. Since the class prior estimation is not our main concern in this paper, the details about 3C are presented in the supplementary materials.

\begin{figure*}[t]
	\centering
	\includegraphics[width=0.8\linewidth]{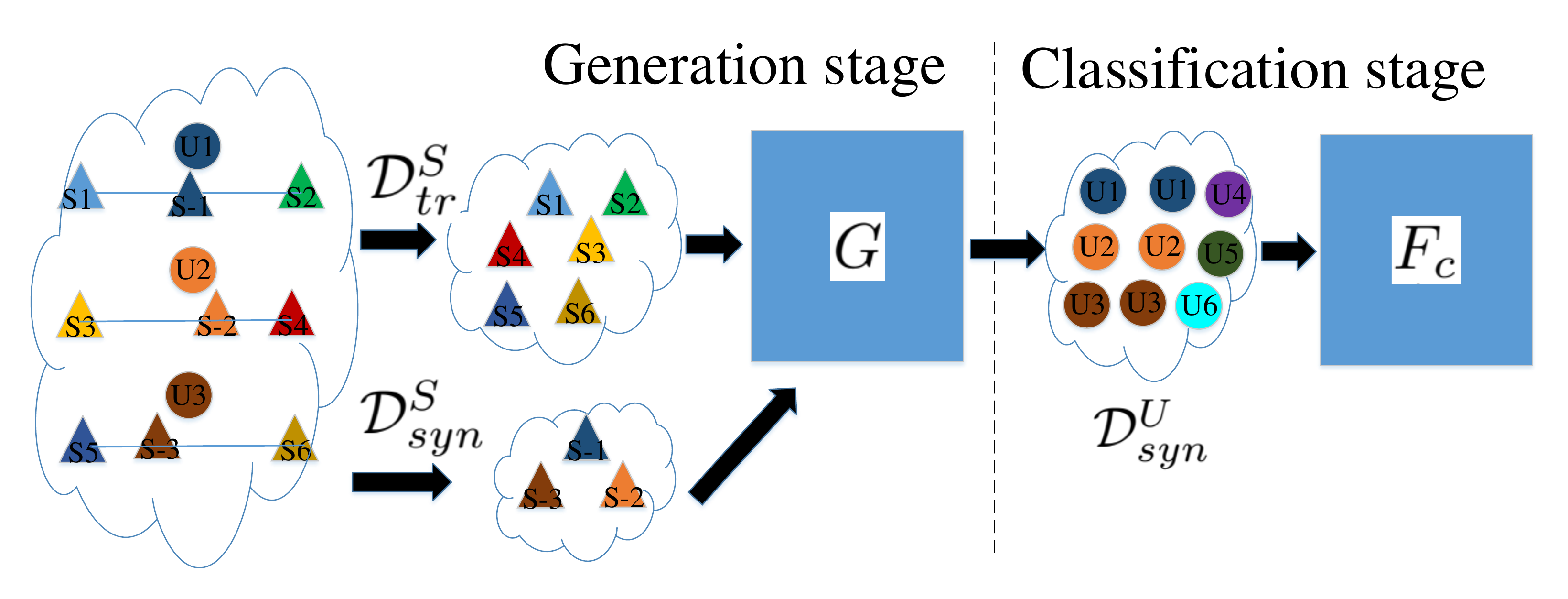}
	\caption{The proposed HarS framework. At the generation stage, some extra `hard classes' (e.g. S-1, S-2, and S-3) are synthesized by interpolating between two seen classes (e.g. S1 with S2, S3 with S4, and S5 with S6, which are semantically close to a hard unseen class respectively, i.e. U1, U2, and U3) to train generator. At the classification stage, more hard-class samples (e.g. U1, U2, and U3) are synthesized to train classifier.}
	\label{fig4}
\end{figure*}

\subsection{How to Boost ZSL with Hard Classes}
In this section, the goal is to study how to boost ZSL with hard classes identified by the proposed metrics in the above section. To this end, two frameworks are proposed here, i.e. HarS for inductive ZSL and HarST for transductive ZSL, which are elaborated below.

\subsubsection{Inductive ZSL Setting---HarS Framework}
Here we propose a \textbf{Har}dness based \textbf{S}ynthesizing framework to boost inductive ZSL by making use of hard-class information. The proposed framework consists of two components: 1) a base inductive ZSL model which could be an arbitrary generative ZSL model, including a generator $G$ and a classifier $F_{c}$; 2) two hard-class sample synthesizing methods. In this paper, our current framework is mainly designed to accommodate and improve generative ZSL methods. This means that our framework also includes the general two stages as done in all generative ZSL methods, where a generative model $G$ is firstly learned with the seen-class data $\mathcal{D}^{S}_{tr}$ to generate unseen-class samples (i.e. the generation stage), and a classifier $F_{c}$ is then trained with these generated unseen-class samples to classify real unseen-class ones (i.e. the classification stage).

The overall procedure of the proposed HarS framework is as follows. At first, the proposed SS metric in Section 3.4.2 is utilized to identify $K$ hard classes among unseen classes. Then, as show in Figure~\ref{fig4}, hard-class information is injected into the proposed HarS in both the generation stage and the classification stage:

\textcircled{1}At the generation stage, the generator $G$ is expected to pay more attention to the `hard classes'. Here the `hard classes' do not refer to the identified $K$ hard (unseen) classes, but some synthesized (hallucinated) `hard classes'. Concretely, around each of the identified $K$ hard unseen classes, we synthesize some `hard class' samples by two steps: 1) firstly finding the Top-$S$ most similar seen classes with the given identified hard unseen class based on semantic similarities, named `support seen classes' for ease of reading,  where $S$ is always set to $2$ in this work; 2) then randomly interpolating between the samples from the `support seen classes' by:
\begin{equation}
	\label{eq5}
	\begin{split}
		x^{s} & = \gamma*x_{i}^{s} + (1-\gamma)*x_{j}^{s}  \\
		e^{s} & = \gamma*e_{i}^{s} + (1-\gamma)*e_{j}^{s}  \\
	\end{split}
\end{equation}
where $x^{s}$ and $e^{s}$ are the synthesized `hard class' samples and semantic features respectively, $x_{i}^{s}$ and $x_{j}^{s}$ are randomly sampled from the `support seen classes', $e_{i}^{s}$ and $e_{j}^{s}$ are the corresponding semantic features to $x_{i}^{s}$ and $x_{j}^{s}$ respectively, and $\gamma$ is sampled from a uniform distribution $\mathcal{U}(0,1)$. The most left part of Figure~\ref{fig4} show this process. After that, a synthesized `hard class' dataset is obtained, denoted by $\mathcal{D}^{S}_{syn}$. Quantitatively, using the samples from the `support seen classes', $\alpha*N_{s}$ `hard class' samples are synthesized, where $N_{s}$ is the total number of the samples from the `support seen classes' and $\alpha$ is a hyper-parameter to control the scale of synthesized `hard class' samples, called the seen synthesizing scale. Finally, $\mathcal{D}^{S}_{syn}$ is combined with the seen-class dataset $\mathcal{D}^{S}_{tr}$ to train the generative model $G$ according to the corresponding objective function of the adopted generative ZSL method. 

\begin{algorithm}[t]
	\caption{HarS}  
	\label{alg1}
	\begin{algorithmic}[1]  
		\Require  
		$\mathcal{D}^{S}_{tr}$, $\mathcal{D}^{U}$, $G$, $F_{c}$;   
		\Ensure 
		Predictions $\hat{Y}$ on $\mathcal{D}^{U}$; 
		\State Identify $K$ hard classes with the SS metric;
		\State Synthesize $\mathcal{D}^{S}_{syn}$ according to Equation~\ref{eq5};
		\State Train $G$ with $\mathcal{D}^{S}_{syn}$ and $\mathcal{D}^{S}_{tr}$;
		\State Synthesize $\mathcal{D}^{U}_{syn}$ with the learned $G$;
		\State Train $F_{c}$ with $\mathcal{D}^{U}_{syn}$ according to Equation~\ref{eq6};
		\State Make predictions on $\mathcal{D}^{U}$ with the learned $F_{c}$ according to Equation~\ref{eq7};\\
		\Return $\hat{Y}$;
	\end{algorithmic}
\end{algorithm}

Here, we would like to provide an analysis about the benefit of synthesizing `hard classes' in the training of generative model. In fact, the interpolating around a given hard unseen class with some similar seen classes could be regarded as hallucinating some virtual classes which are probably similar to real unseen classes. For example, the hallucinated class `S-1' is close to the unseen class `U1' in Figure~\ref{fig4}. Trained with such hallucinated classes, the model's discriminability within this local feature space could be strengthened so that the misclassifications could be reduced when classifying unseen-class samples within this local space.

\textcircled{2}At the classification stage, the classifier is also expected to allocate more attention to hard classes by synthesizing more hard-unseen-class samples than easy-unseen-class ones. As illustrated in Figure~\ref{fig4}, twice of the number of samples from U1, U2 and U3 (assume that they are harder than U4, U5 and U6) are generated to train the classifier $F_{c}$. In particular, suppose $N_{u}$ samples per easy unseen class are generated for classifier training, then we generate $\beta*N_{u} (\beta>1)$ samples per hard unseen class to highlight hard classes, where $\beta$ is a hyper-parameter, called unseen synthesizing scale. The synthesized unseen-class data are denoted by $\mathcal{D}^{U}_{syn}$. Then, the classifier $F_{c}$ is trained with $\mathcal{D}^{U}_{syn}$ according to
\begin{equation}
	\min_{F_{c}} E_{(x,y) \in \mathcal{D}^{U}_{syn}} [\mathcal{C}(F_{c}(x), y)]
	\label{eq6}
\end{equation}
where $x$ and $y$ are samples and corresponding labels in $\mathcal{D}^{U}_{syn}$ and $\mathcal{C}(\cdot)$ is a cross-entropy loss function. After training, the classifier $F_{c}$ is used to predict the labels on the unseen-class dataset $\mathcal{D}^{U}$ by 
\begin{equation}
	\hat{y} = \arg \max_{\hat{y} \in Y^{U}} F_{c}(x)
	\label{eq7}
\end{equation}
where $\hat{y}$ is the predicted labels of the input $x$. The whole predictions on $\mathcal{D}^{U}$ are denoted by $\hat{Y}$. Considering that hard classes have been highlighted in classifier training, the learned classifier is expected to achieve better ability to discriminate hard-class samples. The complete procedure is summarized in Algorithm~\ref{alg1}.

As shown in Figure~\ref{fig4}, the proposed HarS framework improves the traditional generative ZSL by adding two key units: the `hard seen class' synthesizing and the hard unseen class synthesizing. In principle, the proposed HarS framework could be used for any generative inductive ZSL methods to further boost their performances. In Section 4, we employ two representative methods, f-CLSWGAN and TF-VAEGAN, to validate the effectiveness of the proposed framework. The corresponding methods are named HarS-WGAN and HarS-VAEGAN respectively.

\begin{figure}[t]
	\centering
	\includegraphics[width=1\linewidth]{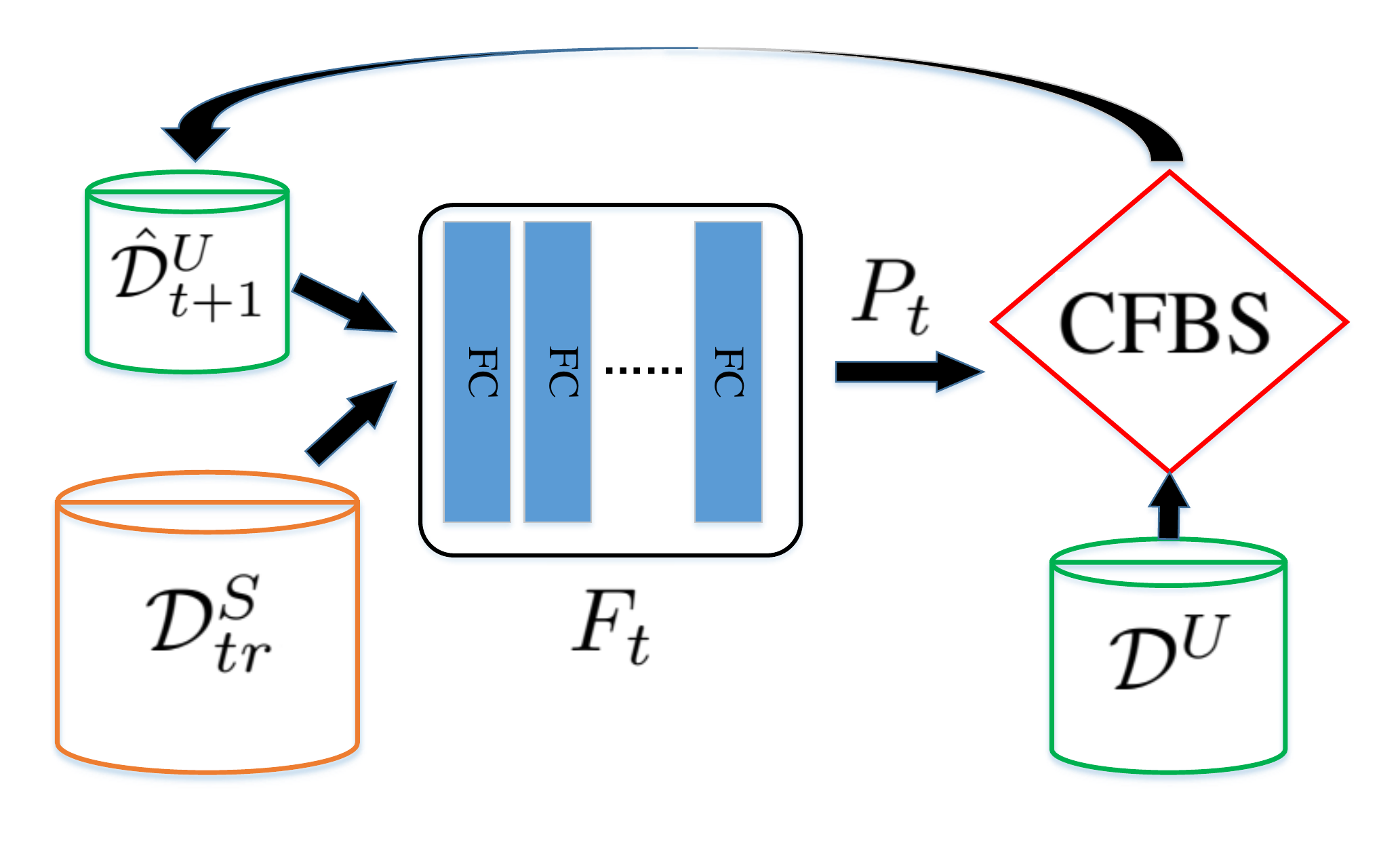}
	\caption{The proposed HarST framework is an iterative framework. At the $t$-th iteration, a set of pseudo labels $P_{t}$ on unseen-class data $\mathcal{D}^{U}$ is firstly predicted by the model $F_{t}$, then a subset of pseudo-labeled unseen-class data $\mathcal{\hat{D}}^{U}_{t+1}$ is selected according to CFBS to update the training set, consisting of $\mathcal{\hat{D}}^{U}_{t+1}$ and $\mathcal{D}^{S}_{tr}$, and used to re-train the model at the $t+1$-th iteration.}
	\label{fig5}
\end{figure}

\begin{algorithm}[t]
	\caption{HarST}  
	\label{alg2}
	\begin{algorithmic}[1]  
		\Require  
		$\mathcal{D}^{S}_{tr}$, $\mathcal{D}^{U}$, $F_{0}$; 
		\Ensure  
		Predictions ${P}_{T}$ on $\mathcal{D}^{U}$; 
		\State Initialization: train the base model $F_{0}$ with $\mathcal{D}^{S}_{tr}$, obtain $\mathcal{D}_{1}$ according to CFBS (or PnCFBS);
		\For{t=1 to T}
		\State Re-train $F_{t}$ with the training set $\mathcal{D}_{t}$;
		\State Make predictions $P_{t}$ on $\mathcal{D}^{U}$ with the learned $F_{t}$;
		\State Update the training set as $\mathcal{D}_{t+1}$ with $P_{t}$ according to CFBS (or PnCFBS);
		\EndFor \\
		\Return ${P}_{T}$;
	\end{algorithmic} 
\end{algorithm}

\subsubsection{Transductive ZSL Setting---HarST Framework}
In the transductive ZSL setting, we propose a \textbf{Har}dness based \textbf{S}electing \textbf{T}ransductive framework to boost ZSL by using hard-class information. Since unlabeled unseen-class samples are available for model training here, HarST adopts a self-training based framework, as show in Figure~\ref{fig5}, mainly including two components: 1) an arbitrary base inductive ZSL model (i.e. $F_{t}$); 2) an unseen-class sample selection module (e.g. CFBS introduced later). The whole framework proceeds in an iterative training manner.

Specifically, assume that we are given a base inductive ZSL model $F_{0}$ and the iterative training process includes $T$ steps. At the $t$-th ($1 \le t \le T$) step, the training set $\mathcal{D}_{t}$ consists of the labeled seen-class dataset $\mathcal{D}^{S}_{tr}$ and a pseudo-labeled unseen-class subset $\mathcal{\hat{D}}^{U}_{t}$. We firstly re-train the model with the training set $\mathcal{D}_{t}$, obtaining the updated model $F_{t}$. Then, the pseudo labels of the whole unseen-class dataset $\mathcal{D}^{U}$ are predicted with the model $F_{t}$, obtaining the pseudo label set $P_{t}$. Then, a new pseudo-labeled unseen-class subset $\mathcal{\hat{D}}^{U}_{t+1}$ is selected from the whole unseen-class dataset $\mathcal{D}^{U}$ by the following two steps: 1) identifying $K$ hard classes with the proposed CF (or PnCF) metric according to Equation~\ref{eq4}; 2) randomly choosing the same number of samples from each of the $K$ hard classes with replacement. For ease of reading, the two steps are named CF (or PnCF) based selection, (CFBS (or PnCFBS) for short). After selecting $\mathcal{\hat{D}}^{U}_{t+1}$, the training set is updated by $\mathcal{D}_{t+1} = \mathcal{D}^{S}_{tr} + \mathcal{\hat{D}}^{U}_{t+1}$ for re-training the model at the $t+1$-th step. Note that uniformly sampling from the $K$ hard classes with replacement is to prevent some classes with small numbers of samples from being overwhelmed by those classes with large numbers of samples. In addition, given the fact that pseudo labels are always more noisy at the beginning, a step-wise incremental learning manner is designed to select the pseudo-labeled unseen-class subset. Specifically, suppose the whole unseen-class dataset $\mathcal{D}^{U}$ contains $M$ samples, then at the $t$-th iteration, $[t*M/(T*K)]$ samples are sampled for each hard class, where $[\cdot]$ is a floor function. The complete procedure is summarized in Algorithm~\ref{alg2}. 

Note that the proposed HarST framework is able to accommodate any inductive ZSL methods. In this work two typical methods, DEM and f-CLSWGAN are embedded into the HarST framework to demonstrate its effectiveness, called HarST-DEM and HarST-WGAN respectively in the next Section.

\section{Experiments}

\subsection{Datasets and Comparative Methods}
The proposed two frameworks are evaluated on three widely used datasets, including AWA2 (Animals with Attributes2~\cite{Xian17Comprehensive}), CUB (Caltech USCD Birds-2011~\cite{WahCUB_200_2011}), and SUN (SUN attributes~\cite{patterson2012sun}). AWA2 is a relatively coarse-grained dataset with totally $37,322$ images from $50$ animal classes and each class is annotated by $85$ attributes. CUB is a middle-scale fine-grained bird dataset, which contains $11,788$ images from $200$ bird species and each bird class is annotated by $312$ attributes. SUN is a relatively large-scale fine-grained dataset consisting of $14,340$ images belonging to $717$ scene categories, where each class is annotated by $102$ attributes. Following most existing works~\cite{Xian17Comprehensive}, visual features extracted from the original images in these datasets by an ImageNet1000~\cite{deng2009imagenet} pre-trained ResNet101~\cite{He16resnet} are directly used as the model's inputs in this paper and attributes are used as semantic features accordingly. As for data split, since some unseen classes in these datasets according to the standard split (SS) are actually included in ImageNet1000, which does not strictly conform to the ZSL standard, we evaluate the proposed frameworks with the PS split~\cite{Xian17Comprehensive} for fair evaluation and comparison. In PS, $40/150/645$ classes are regarded as seen classes in AWA2/CUB/SUN respectively, while the remaining $10/50/72$ classes are taken as unseen classes accordingly.

The proposed two HarS based inductive methods, HarS-WGAN and HarS-VAEGAN are compared with $18$ state-of-the-art inductive methods: DEVISE~\cite{Frome13DeViSE}, LATEM~\cite{xian2016LATEM}, SAE~\cite{Kodirov17SAE}, DEM~\cite{zhang2017DEM}, LiGAN~\cite{li2019LiGAN}, ABP~\cite{zhu2019ABP}, TCN~\cite{jiang2019TCN}, OCD-GZSL~\cite{keshari2020OCDZSL}, APNet~\cite{liu2020APNet}, DE-VAE~\cite{Ma2020DE-VAE}, LsrGAN~\cite{Vyas2020LrGAN}, f-CSLWGAN~\cite{Xian18FCLSWGAN}, DASCN~\cite{NiZ019dascn}, DVBE~\cite{dvbe2020}, DAZLE~\cite{huynh2020DAZLE}, TF-VAEGAN~\cite{Narayan2020TF-VAEGAN}, CF-GZSL~\cite{Han_2021_CVPR}, GCM-CF~\cite{yue2021counterfactual}. The proposed two HarST based transductive
methods, HarST-DEM and HarST-WGAN are compared with $14$ state-of-the-art transductive methods: ALE-tran~\cite{Akata16ALE}, GFZSL~\cite{verma2017GFZSL}, DSRL~\cite{Ye2017DSRL}, QFSL~\cite{Song18QFSL}, GMN~\cite{sariyildiz2019GMN}, f-VAEGAN-D2~\cite{xian2019f-VAEGAN}, GXE~\cite{li2019GXE}, SABR-T~\cite{paul2019SABR}, PREN~\cite{YeG19PREN}, VSC~\cite{wan2019vsc}, DTN~\cite{Zhang2020DeepTN}, ADA~\cite{Khare2020ADA}, SDGN~\cite{wu2020SDGN}, TF-VAEGAN~\cite{Narayan2020TF-VAEGAN}.

\subsection{Evaluation Protocol}
As done in the most existing works~\cite{Xian17Comprehensive}, we evaluate the proposed methods in both ZSL and GZSL settings. In the ZSL setting, the performances are evaluated by computing average per-class Top-1 accuracy (ACC) on unseen classes, denoted as $ACC_{U}$. In the GZSL setting, the performances are evaluated by firstly computing ACC on unseen classes and seen classes and then computing their harmonic mean, denoted as $ACC_{U}$, $ACC_{S}$, $H$ respectively, where
\begin{equation}
	H = \frac{2*ACC_{U}*ACC_{S}}{ACC_{U}+ACC_{S}}
\end{equation}

\subsection{Implementation Details}
In the HarS framework, f-CLSWGAN~\cite{Xian18FCLSWGAN} and TF-VAEGAN~\cite{Narayan2020TF-VAEGAN} are employed as the base models. The seen synthesizing scale $\alpha$ and the unseen synthesizing scale $\beta$ are set to $2$ in AWA2/CUB/SUN for both f-CLSWGAN and TF-VAEGAN. In the HarST framework, DEM~\cite{zhang2017DEM} and f-CLSWGAN are employed as the base models. The iteration number ($T$) is set as $6/9/12$ and $5/6/4$ for DEM and f-CLSWGAN on AWA2/CUB/SUN respectively. The network architectures of TF-VAEGAN and DEM are totally consistent with those in the original papers. The architecture of f-CLSWGAN is slightly modified from the original one for ease of implementation. The details about these network architectures are described in the supplementary materials since they are not our main concerns in this paper. In addition, the training configurations of the three base models are generally consistent with those in the original papers. Specifically, DEM is trained with $5/30/30$ epochs on AWA2/CUB/SUN respectively, and the batch size and learning rate are set to $128$ and $0.001$ respectively for all the three datasets. The generative model of f-CLSWGAN is trained by $20/50/50$ epochs on AWA2/CUB/SUN, respectively, with the same learning rate of $0.0002$ and same batch size of $128$. The classifier in f-CLSWGAN is trained by $30$ epochs with learning rate of $0.0002$ and batch size of $256$ on all the three datasets. To train the generator of TF-VAEGAN, the training epochs and learning rate are set as $120/200/300$ and  $0.00001/0.0001/0.0005$ on AWA2/CUB/SUN, respectively, and the batch size is $64$ on all the three datasets. For training the classifier in TF-VAEGAN, the training epochs, batch size and learning rate are set as $25$, $64$, and $0.0005$ on all the three datasets respectively.

\subsection{Improvements in the ZSL Setting}
Here we evaluate the proposed two frameworks in the ZSL tasks on AWA2, CUB and SUN with the PS data split and compare them with corresponding inductive (or transductive) methods. The proposed HarS-WGAN and HarS-VAEGAN are evaluated in the inductive ZSL setting, and the proposed HarST-DEM and HarST-WGAN are evaluated in the transductive setting. Note that the results of the comparative methods are directly cited from the original papers.

The inductive ZSL results are reported in Table~\ref{tab1}. As seen from Table~\ref{tab1}, the performances of both f-CLSWGAN and TF-VAEGAN have been significantly improved by the proposed HarS framework. Specifically, the improvements for f-CLSWGAN are $10.7\%$, $3.9\%$ and $1.8\%$ on AWA2, CUB, and SUN respectively, and those for TF-VAEGAN are $6.1\%$, $3.2\%$ and $0.6\%$. These improvements are substantial even though they cost marginal efforts, which demonstrates that the proposed HarS is quite effective. The large improvements also indicate that hard-class information has a huge potential to boost ZSL, which is worthy of further exploration and more efforts in the future. On the other side, we find from Table~\ref{tab1} that the improvements for both methods (i.e. f-CLSWGAN and TF-VAEGAN) on AWA2 are larger than those on CUB and especially SUN. This is due to the fact that SUN and CUB are relatively large-scale fine-grained datasets, for instance, CUB contains $200$ fine-grained species of birds, and SUN contains $717$ fine-grained kinds of scenes, where semantic relations between classes are less distinct so that hard classes are much more difficult to identify. While AWA2 is a relatively coarse-grained dataset with $50$ different categories of animals. Hence, the performance of identifying hard classes is better in AWA2 than in CUB and SUN. We will quantitatively demonstrate this point later in Section~\ref{analysis}. Lastly, when compared with recent state-of-the-art methods, the proposed HarS-VAEGAN can outperform them with large margins, e.g., $6.7\%$ on AWA2, $3.2\%$ on CUB are achieved respectively.

Table~\ref{tab2} shows the results in the transductive ZSL setting. From Table~\ref{tab2}, we can see that: 1) the proposed HarST is able to improve the performances of existing ZSL methods by significant margins, e.g., the improvements on AWA2 reaches $24.3\%$ and $26.7\%$ for DEM and f-CLSWGAN respectively, which demonstrates that the proposed HarST can select informative pseudo-labeled unseen-class samples to effectively improve inductive ZSL methods by transductive training; 2) HarST-WGAN achieves better performances than recent state-of-the-art methods, e.g., the improvement on CUB reaches $2.5\%$, and HarST-DEM also achieves comparable performances with the state-of-the-art methods; 3) the improvements achieved by the proposed HarST vary across different inductive base ZSL models and different datasets. For instance, DEM receives more improvements than f-CLSWGAN on SUN while the improvement of f-CLSWGAN is larger than that of DEM on CUB. The reason of such variations is that different inductive ZSL models have different performances on different datasets, which then affects the quality of their pseudo labels and the hard class identifications.

\begin{table}[t]
	\centering
	\caption{Comparative results ($ACC_{U}$) in the inductive ZSL setting on AWA2, CUB and SUN.}
	\resizebox{.7\columnwidth}{!}{
		\begin{tabular}{cccc}
			\toprule
			Method&   AWA2& CUB& SUN \\
			\cmidrule{1-4}
			DEVISE&    59.7& 52.0& 56.5 \\
			LATEM&     55.8& 49.3& 55.3 \\
			SAE&       54.1& 33.3& 40.3 \\
			DEM&       67.1& 51.7& 61.9 \\
			LiGAN&     -& 58.8& 61.7 \\
			ABP&       70.4& 58.5& 61.5 \\
			TCN&       71.2& 59.5& 61.5 \\
			OCD-GZSL&  71.3& 60.3& 63.5\\
			APNet&     68.0& 57.7& 62.3 \\
			DE-VAE&    69.3& 63.1& 64.0 \\
			LsrGAN&    	  -& 60.3& 62.5 \\
			CE-GZSL&   70.4&    -& 63.3  \\
			\cmidrule{1-4}
			f-CLSWGAN& 68.2& 57.3& 60.8 \\
			TF-VAEGAN& 72.2& 64.9& 66.0 \\
			\cmidrule{1-4}
			HarS-WGAN(Ours)&  \textbf{78.9}&          61.2&         62.6 \\
			HarS-VAEGAN(Ours)&   	  78.3& \textbf{68.1}& \textbf{66.6} \\
			\bottomrule
		\end{tabular}
	}
	\label{tab1}
\end{table}

\subsection{Improvements in the GZSL Setting}
Here we evaluate the GZSL performances of the proposed two frameworks on AWA2, CUB, and SUN with the PS split. The inductive results of the proposed HarS-WGAN, HarS-VAEGAN and the reference methods are reported in Table~\ref{tab3}. The first observation from Table~\ref{tab3} is that the proposed HarS is able to improve the two base models (i.e. f-CLSWGAN and TF-VAEGAN) with huge margins. Particularly, the improvements on AWA2 achieve $11.6\%$ and $5.4\%$ for f-CLSWGAN and TF-VAEGAN respectively. These significant improvements not only validate the effectiveness of the proposed hard class synthesizing framework but also reveal the importance of hard-class information to ZSL. Secondly, we also observe a similar phenomenon as in the inductive ZSL evaluation that the proposed HarS improves the performances more significantly in the relatively coarse-grained AWA2 than in the relatively fine-grained CUB and SUN. The underlying reason is similar as in ZSL, i.e., hard classes are relatively difficult to accurately identify in the fine-grained large-scale datasets where semantic affinities between classes are less distinct. We will leave this problem as a future work. Last but not least, as shown in Table~\ref{tab3}, the proposed HarS-VAEGAN achieves the new state-of-the-art performances, outperforming the previous one~\cite{Han_2021_CVPR} by $2.0\%$ and $0.4\%$ on AWA2 and SUN respectively.

Among the existing transductive GZSL methods, there are two different ways to use the unlabeled data (including the unlabeled unseen-class data $\mathcal{D}^{U}$ and the unlabeled seen-class data $\mathcal{D}^{S}$). In the first one, $\mathcal{D}^{U}$ is separately used to train the model, which means that $\mathcal{D}^{U}$ and $\mathcal{D}^{S}$ are assumed to be separable at the training stage, and also allows models to predict pseudo labels in the unseen-class space $Y^{U}$. In the second one, $\mathcal{D}^{U}$ and $\mathcal{D}^{S}$ are compound and not separable in the training. In other words, pseudo labels have to be predicted in the total-class space $Y$ instead, meaning that it is a harder setting than the first one. For fair comparison, we adapt the proposed methods to both settings, and compare them with the corresponding methods trained in the same setting.
\begin{table}[t]
	\centering
	\caption{Comparative results ($ACC_{U}$) in the transductive ZSL setting on AWA2, CUB and SUN.}
	\resizebox{.7\columnwidth}{!}{
		\begin{tabular}{cccc}
			\toprule
			Method&   AWA2& CUB& SUN \\
			\cmidrule{1-4}
			ALE-tran&    70.7& 54.5& 55.7 \\
			GFZSL&       78.6& 50.0& 64.0 \\
			DSRL&        72.8& 48.7& 56.8 \\
			QFSL&        79.7& 72.1& 58.3 \\
			GMN&         -& 64.6& 64.3 \\
			f-VAEGAN-D2&  89.8& 71.1& 70.1 \\
			GXE&         83.2& 61.3& 63.5 \\
			SABR-T&      88.9& 74.0& 67.5 \\
			PREN&        78.6& 66.4& 62.8 \\
			VSC&         81.7& 71.0& 62.2 \\
			ADA&         78.6& -& 65.5 \\
			SDGN&        93.4& 74.9& 68.4 \\
			TF-VAEGAN&   92.1& 74.7& \textbf{70.9} \\
			\cmidrule{1-4}
			DEM&        67.1& 51.7& 61.9 \\
			f-CLSWGAN&  68.2& 57.3& 60.8 \\
			\cmidrule{1-4}
			HarST-DEM(Ours)&            91.4&          71.2& \textbf{70.9} \\
			HarST-WGAN(Ours)&  \textbf{94.9}& \textbf{77.4}&          67.5 \\
			\bottomrule
		\end{tabular}
	}
	\label{tab2}
\end{table}

\begin{table*}[t]
	\centering
	\caption{Comparative results in the inductive GZSL setting on AWA2, CUB and SUN.}
	\resizebox{1.5\columnwidth}{!}{
		\begin{tabular}{cccccccccc}
			\toprule
			Method& \multicolumn{3}{c}{AWA2}& \multicolumn{3}{c}{CUB}& \multicolumn{3}{c}{SUN} \\
			\cmidrule(lr){2-4} \cmidrule(lr){5-7} \cmidrule(lr){8-10}
			&  $ACC_{U}$&  $ACC_{S}$&  $H$& $ACC_{U}$&  $ACC_{S}$&  $H$&  $ACC_{U}$&  $ACC_{S}$&  $H$ \\
			\cmidrule{1-10}
			DASCN&          -&     -&   -& 45.9& 59.0& 51.6& 42.4& 38.5& 40.3 \\
			DVBE&        63.6& 70.8& 67.0& 53.2& 60.2& 56.5& 45.0& 37.2& 40.7 \\
			DAZLE&       60.3& 75.7& 67.1& 56.7& 59.6& 58.1& 52.3& 24.3& 33.2 \\
			OCD-GZSL&    59.5& 73.4& 65.7& 44.8& 59.9& 51.3& 44.8& 42.9& 43.8 \\
			APNet&       54.8& 83.9& 66.4& 48.1& 55.9& 51.7& 35.4& 40.6& 37.8 \\
			DE-VAE&      58.8& 78.9& 67.4& 52.5& 56.3& 54.3& 45.9& 36.9& 40.9 \\
			LsrGAN&         -&    -&    -& 48.1& 59.1& 53.0& 44.8& 37.7& 40.9 \\
			CE-GZSL&	 63.1& 78.6& 70.0& 	  -&    -&    -& 48.8& 38.6& 43.1 \\
			GCM-CF&	 	 60.4& 75.1& 67.0& 61.0& 59.7& \textbf{60.3}& 47.9& 37.8& 42.2 \\
			\cmidrule{1-10}
			f-CLSWGAN&   57.9& 61.4& 59.6& 43.7& 57.7& 49.7& 42.6& 36.6& 39.4 \\
			TF-VAEGAN&   59.8& 75.1& 66.6& 52.8& 64.7& 58.1& 45.6& 40.7& 43.0 \\
			\cmidrule{1-10}
			HarS-WGAN(Ours)&     65.8& 77.4& 		 71.2&  49.1& 57.6& 	     53.0& 45.5& 37.1&         40.9 \\
			HarS-VAEGAN(Ours)&   64.1& 81.2& \textbf{72.0}&  54.8& 63.6& 58.9& 49.5& 38.8& \textbf{43.5} \\
			\bottomrule
		\end{tabular}
	}
	\label{tab3}
\end{table*}
In the first setting, since $\mathcal{D}^{U}$ can be separately used for model training, we adapt the proposed HarST-DEM and HarST-WGAN as follows. Firstly, we train an OOD detector with $\mathcal{D}^{U}$ and the labeled seen-class dataset $\mathcal{D}^{S}_{tr}$ according to a baseline OOD detection method~\cite{Hendrycks19OE}. Since OOD detection is out of the scope of this paper, we put its descriptions in the supplementary materials. Then, we train a seen-class classifier with $\mathcal{D}^{S}_{tr}$ according to a cross-entropy loss function, and an unseen-class classifier using the proposed ZSL methods (i.e. HarST-DEM and HarST-WGAN). At the testing stage, given the testing data, we firstly classify them into seen/unseen classes with the OOD detector, and then classify the `seen-class' data with the seen-class classifier and `unseen-class' data with the unseen-class classifier. The corresponding results are reported at the top of Table~\ref{tab4}. The results show that the proposed two HarST based methods outperform recent state-of-the-art methods with significant margins, e.g., the improvements about $4.5\%$, $4.9\%$, and $0.2\%$ on AWA2, CUB, and SUN are achieved by HarST-WGAN. We also note that the improvement on SUN is relatively smaller than those on AWA2 and CUB. This is because the OOD detection on SUN is considerably harder considering that SUN contains $717$ fine-grained categories with only $20$ samples per class.

In the second setting, since $\mathcal{D}^{U}$ and $\mathcal{D}^{S}$ are compound in the training, models have to predict pseudo labels in the total-class space $Y$, i.e. the pseudo-label prediction is actually a GZSL task. Considering that embedding based ZSL methods usually suffer severely from the bias problem~\cite{socher2013CMT} in GZSL and generative methods are able to deal with it to some extent, here we validate the proposed HarST framework by evaluating HarST-WGAN. The corresponding results are reported at the bottom of Table~\ref{tab4}. The results show that the performances in the second setting are generally worse than those in the first setting, which is expected because the pseudo labels are harder to predict in the second setting. In addition, HarST-WGAN outperforms its base ZSL model (i.e. f-CLSWGAN) significantly, e.g., the improvements reach $23.6\%$ and $16.5\%$ on AWA2 and CUB, respectively. The huge improvements suggest that the proposed HarST is effective to make use of the unlabeled data in the GZSL setting to improve the performance. In contrast, the improvement on SUN is limited, since pseudo label prediction on SUN with $717$ fine-grained classes is quite difficult. In addition, HarST-WGAN achieves significantly superior performances over the competitors on all three datasets. This is partly because a generative ZSL model (i.e. f-CLSWGAN) which is good at the GZSL task is employed as the base model in HarST-WGAN, but more importantly, this is because the proposed HarST can successfully select hard-class samples to further boost the GZSL performances.

\begin{table*}[t]
	\centering
	\caption{Comparative results in the transductive GZSL setting on AWA2, CUB and SUN.}
	\resizebox{1.5\columnwidth}{!}{
		\begin{tabular}{cccccccccc}
			\toprule
			Method& \multicolumn{3}{c}{AWA2}& \multicolumn{3}{c}{CUB}& \multicolumn{3}{c}{SUN} \\
			\cmidrule(lr){2-4} \cmidrule(lr){5-7} \cmidrule(lr){8-10}
			&  $ACC_{U}$&  $ACC_{S}$&  $H$& $ACC_{U}$&  $ACC_{S}$&  $H$&  $ACC_{U}$&  $ACC_{S}$&  $H$ \\
			\cmidrule{1-10}
			f-VAEGAN-D2&    84.8& 88.6& 86.7& 61.4& 65.1& 63.2& 60.6& 41.9& 49.6 \\
			GMN&               -&    -&    -& 60.2& 70.6& 65.0& 57.1& 40.7& 47.5 \\
			SABR-T&         79.7& 91.0& 85.0& 67.2& 73.7& 70.3& 58.8& 41.5& 48.6 \\
			GXE&            80.2& 90.0& 84.8& 57.0& 68.7& 62.3& 45.4& 58.1& 51.0 \\
			VSC&            71.9& 88.2& 79.2& 33.1& 86.1& 47.9& 29.9& 62.9& 40.6 \\
			SDGN&           88.8& 89.3& 89.1& 69.9& 70.2& 70.1& 62.0& 46.0& 52.8 \\
			TF-VAEGAN&      87.3& 89.6& 88.4& 69.9& 72.1& 71.0& 62.4& 47.1& 53.7 \\
			\cmidrule{1-10}
			HarST-DEM+OD(Ours)&      91.4& 92.3& 91.8&  71.2& 74.5& 72.8& 70.7& 44.8& \textbf{54.8} \\
			HarST-WGAN+OD(Ours)&     94.9& 92.3& \textbf{93.6}&  77.4& 74.5& \textbf{75.9}& 67.5& 44.8& 53.9 \\
			\cmidrule{1-10}
			\cmidrule{1-10}
			ALE-tran&   12.6& 73.0& 21.5& 23.5& 45.1& 30.9& 19.9& 22.6& 21.2 \\
			GFZSL&     	   -&    -&    -& 24.9& 45.8& 32.2&    -&    -& - \\
			DSRL&          -&    -&    -& 17.3& 39.0& 24.0& 17.7& 25.0& 20.7 \\
			PREN&       32.4& 88.6& 47.4& 35.2& 55.8& 43.1& 35.4& 27.2& 30.8 \\
			DTN&     	-&    -&    -&  42.6& 66.0& 51.8&    35.8&    38.7& 37.2 \\
			\cmidrule{1-10}
			f-CLSWGAN&   57.9& 61.4& 59.6& 43.7& 57.7& 49.7& 42.6& 36.6& 39.4 \\
			\cmidrule{1-10}
			HarST-WGAN(Ours)&  85.4& 81.0& \textbf{83.2}&  65.1& 67.4& \textbf{66.2}& 45.1& 36.2& \textbf{40.2} \\
			\bottomrule      
		\end{tabular}  
	}
	\label{tab4}
\end{table*}

\subsection{Analysis and Discussion}
\label{analysis}

\subsubsection{Hard Class Identification}
Here we verify whether the proposed two metrics, SS and CF, are able to identify hard classes in both the inductive and transductive ZSL settings. All the experiments are conducted on AWA2, CUB, and SUN. 

In the inductive setting, since hard classes are static, we only need to identify them once as follows. Firstly, we compute the true hard (and easy) classes by firstly ranking unseen classes according to their per-class accuracies in the ascending order and then taking the front half and the back half as true hard classes and true easy classes respectively, i.e. there are $5/5$, $25/25$, and $36/36$ easy/hard classes in AWA2, CUB, and SUN respectively. Then, we compute the SS metric ($d$) of each unseen class and obtain the predicted hard (and easy) classes according to their per-class $d$, specifically, the half of unseen classes with the smallest $d$ are predicted as hard classes, the remaining are easy classes. Finally, we compute the average per-class recall (APR) of hard classes according to the true hard classes and the predicted ones. The results on AWA2, CUB, and SUN are $5/5$, $19/25$, and $27/36$ respectively, which demonstrates the effectiveness of the proposed SS metric for hard class identification in the inductive setting. In addition, it also indicates that hard class identification on the relatively coarse-grained AWA2 is better than those on the relatively fine-grained and large-scale CUB and SUN.

\begin{figure}[t]
	\centering
	\includegraphics[width=1\linewidth]{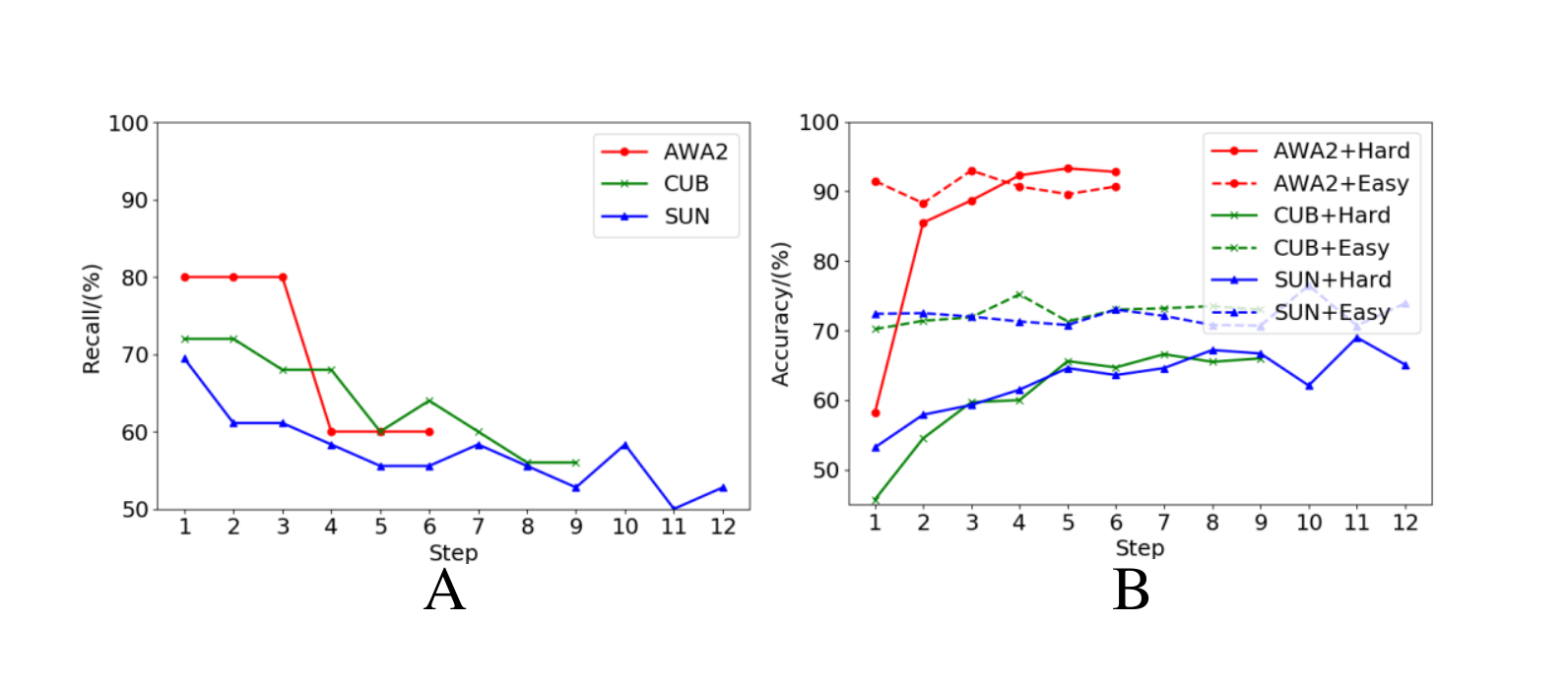}
	\caption{A: APR of unseen classes on AWA2, CUB, and SUN. B: APA of the predicted hard-class (and easy-class) samples on AWA2, CUB, and SUN.}
	\label{fig6}
\end{figure}

In the transductive setting, considering that the predicted hard classes vary during the iterative learning process, we record the identified hard classes with HarST-DEM and HarST-WGAN at each iteration and evaluate their performances by two approaches. Specifically, at each iteration, we first compute the true hard (and easy) classes according to their per-class accuracies as done above. Meanwhile, we also predict the hard (and easy) classes according to the proposed CF (or PnCF) metric in a way similar to that in the inductive setting. Note that here AWA2 is used to simulate the extremely unbalanced dataset, hence PnCF is used on AWA2 while CF is used on CUB and SUN. After obtaining the true hard classes and the predicted ones, in the first evaluation, we compute the average per-class recall (APR) of hard classes. The corresponding results are shown in Figure~\ref{fig6}-A. In the second evaluation, we compute the average per-class accuracy (APA) of the predicted hard-class samples and that of the predicted easy-class samples respectively. By comparing their APA, we can assess whether the predicted hard classes by our metrics are really the relatively hard ones among all unseen classes. The results are shown in Figure~\ref{fig6}-B. Since the results of HarST-WGAN are similar with those of HarST-DEM, we report them in the supplementary materials due to the limited space. 

From Figure~\ref{fig6}-A, we can see that the proposed metrics have considerable ability to identify the true hard classes, especially at the initial stages. With the iterative training going on, the number of real hard classes is actually reduced since accuracies of every classes become considerably high (especially in the AWA2, the identified hard classes even have slightly better accuracy than the easy ones as shown in Figure~\ref{fig6}-B), hence the hard class identification becomes less necessary and their performances are also decreased. Figure~\ref{fig6}-B shows that the APA of the identified hard-class samples is significantly worse than that of the identified easy-class ones in most cases, which indicates that our proposed CF metric can effectively find the relatively hard classes among all unseen classes. In sum, the results in both evaluations demonstrate that the proposed CF metric is able to dynamically identify hard classes among all unseen classes in the transductive iterative training process.

\subsubsection{Pseudo Label Quality of Identified Hard Classes}
\begin{figure}[t]
	\centering
	\includegraphics[width=0.55\linewidth]{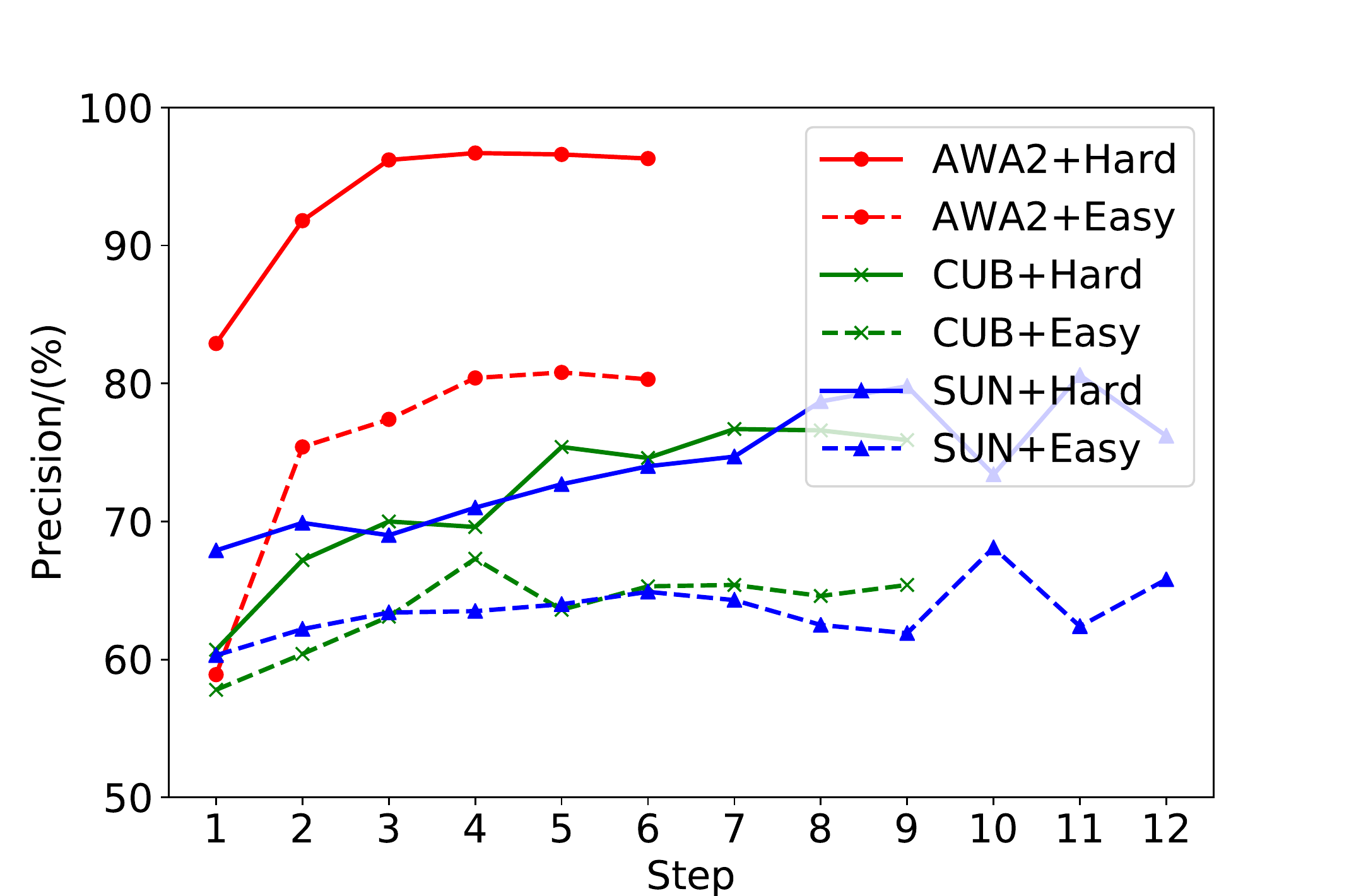}
	\caption{APP of the pseudo labels of the predicted hard-class (and easy-class) samples on AWA2, CUB, and SUN.}
	\label{fig7}
\end{figure}

In this section, we investigate whether the quality of the pseudo labels of the predicted hard classes is better than that of the predicted easy classes in the transductive iterative training process. The experiments are conducted on AWA2, CUB and SUN in the ZSL tasks with HarST-DEM and HarST-WGAN. Specifically, at each iteration, we firstly predict the hard (and easy) classes according to our proposed CF (or PnCF) metric as done in the above section. Then, we compute the average per-class precision (APP) of the identified hard-class samples and that of the identified easy-class ones. The results of HarST-DEM at each iteration are shown in Figure~\ref{fig7}. Figure~\ref{fig7} shows that the hard classes identified by the proposed metrics have significantly higher quality of pseudo labels than the identified easy ones. As the results of HarST-WGAN are similar to those of HarST-DEM, they are put in the supplementary materials due to the limited space.

\subsubsection{Effect of Hardness Based Synthesizing}
Here we analyze the effects of the proposed hardness based synthesizing (HarS) framework on the ZSL performance by conducting experiments on AWA2, CUB and SUN in the inductive ZSL setting with HarS-WGAN and HarS-VAEGAN. In the proposed HarS framework, hard-seen-class synthesizing (HSS) and hard-unseen-class synthesizing (HUS) are two crucial components, which are used at the generation stage and the classification stage separately. Hence, their effects were investigated individually by ablation study. The results ($ACC_{U}$) of HarS-WGAN and HarS-VAEGAN with (or without) HSS and HUS are summarized in Table~\ref{tab5}. From Table~\ref{tab5}, we can see that 1) both HSS and HUS are able to boost the performances significantly on all the three datasets for both HarS-WGAN and HarS-VAEGAN; 2) HSS and HUS have different effects on the ZSL methods on different datasets. This is because the performances of hard class identification on different datasets are different, which affects HSS and HUS to some degree.
\begin{table}[t]
	\centering
	\caption{Comparative results with (or without) HSS and HUS on AWA2, CUB and SUN.}
	\resizebox{0.8\columnwidth}{!}{
		\begin{tabular}{ccccccc}
			\toprule
			Method&  \multicolumn{3}{c}{HarS-WGAN}& \multicolumn{3}{c}{HarS-VAEGAN} \\
			\cmidrule(lr){2-4} \cmidrule(lr){5-7}
			&  AWA2&  CUB&  SUN&  AWA2&  CUB&  SUN \\
			\cmidrule{1-7}
			None&  		68.2& 57.3& 60.8& 72.2& 64.9& 66.0 \\
			HSS&  		76.3& 60.5& 61.7& 76.9& 67.8& 66.5 \\
			HUS&  		76.6& 61.2& 62.4& 76.8& 68.1& 66.4 \\
			HSS+HUS&    78.9& 61.2& 62.6& 78.3& 68.1& 66.6 \\
			\bottomrule
		\end{tabular}
	}
	\label{tab5}
\end{table}

\subsubsection{Effect of Hardness Based Selecting}
Here we demonstrate the effectiveness of the proposed hardness based selecting transductive (HarST) framework in the transductive ZSL setting. The experiments are conducted on AWA2, CUB and SUN in the ZSL tasks with HarST-DEM and HarST-WGAN. In the proposed HarST, two CF base selection methods (i.e CFBS and PnCFBS) are designed to select a subset of pseudo-labeled unseen-class data for model iterative training. For comparison, we also adopt a random selection (RS) based method to carry out this task in the iterative training process. The results are reported in Table~\ref{tab6}. From Table~\ref{tab6}, we find firstly that the performances with CFBS (or PnCFBS) are better than those with RS by large margins. Secondly, it shows that PnCFBS can outperform CFBS when the dataset is extremely unbalanced, i.e. AWA2. Meanwhile, it also shows that PnCFBS performs slightly worse than CFBS on CUB and SUN. This phenomenon could be explained from two aspects. On the one hand, CFBS is able to deal with some relatively unbalanced datasets. On the other hand, PnCFBS is affected by the performance of class prior estimation, especially when faced with a dataset which has large number of classes and small number of per-class samples. Anyway, the proposed HarST with either CFBS or PnCFBS is better than the RS based one and it is also able to significantly improve the existing ZSL methods.
\begin{table}[t]
	\centering
	\caption{Comparative results with different selection methods in HarST on AWA2, CUB and SUN.}
	\resizebox{0.8\columnwidth}{!}{
		\begin{tabular}{ccccccc}
			\toprule
			Method&  \multicolumn{3}{c}{HarST-DEM}& \multicolumn{3}{c}{HarST-WGAN} \\
			\cmidrule(lr){2-4} \cmidrule(lr){5-7}
			&  AWA2&  CUB&  SUN&  AWA2&  CUB&  SUN \\
			\cmidrule{1-7}
			RS&  85.0& 66.0& 66.7& 87.5& 73.1& 63.1 \\
			CFBS&  88.9& 71.2& 70.9& 89.8& 77.4& 67.5 \\
			PN-CFBS&  91.4& 70.0& 70.0& 94.9& 74.5& 65.5 \\
			\bottomrule
		\end{tabular}
	}
	\label{tab6}
\end{table}

\subsubsection{Hyper-Parameter Sensitivity}
\begin{figure}[t]
	\centering
	\includegraphics[width=1\linewidth]{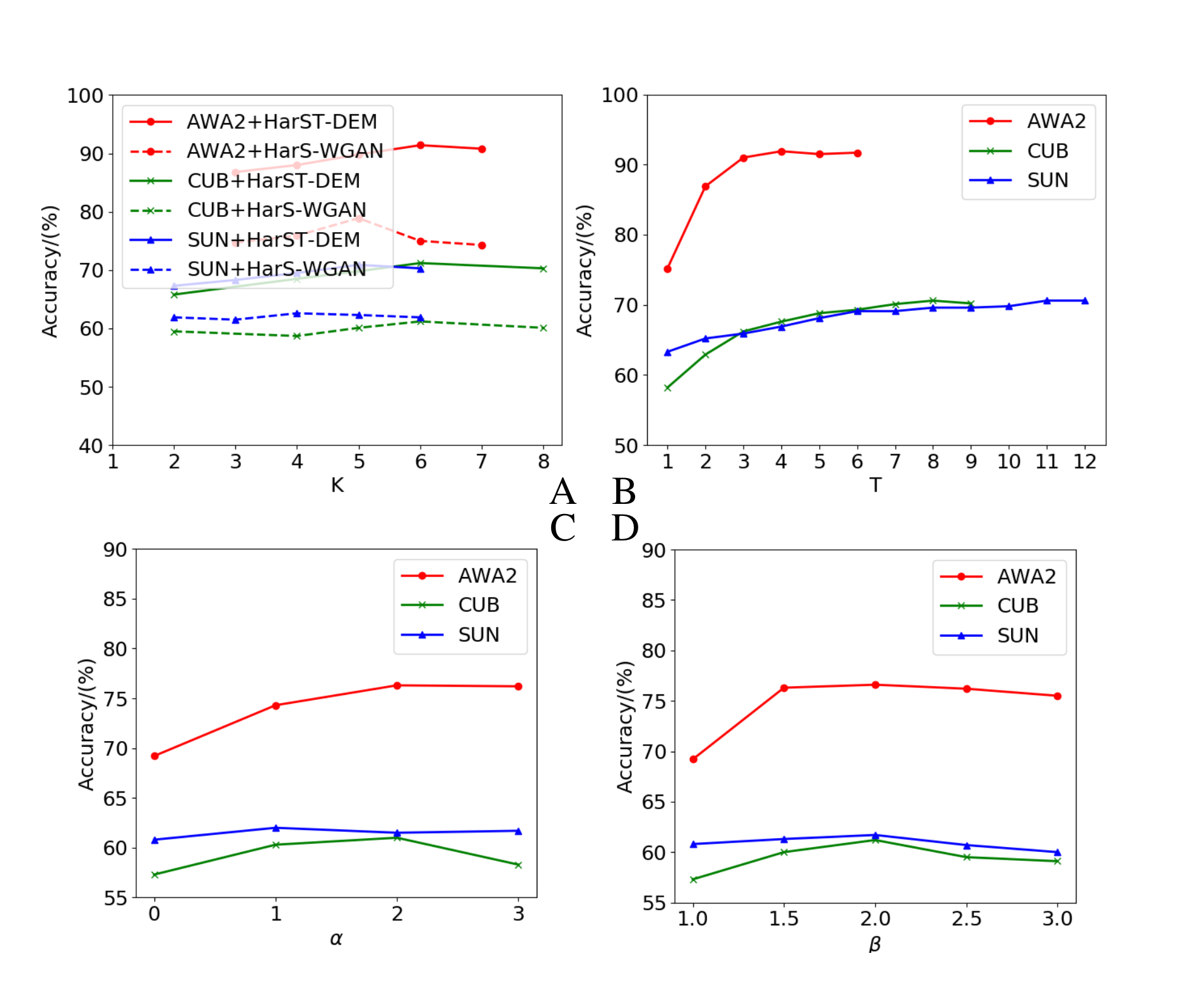}
	\caption{The sensitivity of ZSL performance to the number of hard classes ($K$), the number of iterations ($T$), the seen synthesizing scale ($\alpha$), and the unseen synthesizing scale ($\beta$).}
	\label{fig8}
\end{figure}

Here we analyze the sensitivity of the proposed two frameworks' performances to some hyper-parameters, including the number of the identified hard classes ($K$), the number of iterations ($T$), the seen synthesizing scale ($\alpha$), and the unseen synthesizing scale ($\beta$). We evaluate $K$ in the inductive setting with HarS-WGAN and in the transductive setting with HarST-DEM. $T$ is evaluated in the transductive setting with HarST-DEM. $\alpha$ and $\beta$ are evaluated in the inductive setting with HarS-WGAN. All the experiments are conducted on AWA2, CUB and SUN in the ZSL tasks. The varying ranges of these hyper-parameters and the corresponding results ($ACC_{U}$) are shown in Figure~\ref{fig8}. From Figure~\ref{fig8}, we can see that the performances of the proposed frameworks are generally not very sensitive to these hyper-parameters. It can also be found that 1) choosing an appropriate number of classes as hard classes is beneficial to the final performances. In other words, choosing too few classes is not able to highlight hard classes while choosing too many classes would lessen the benefit of real hard classes; 2) with the iterative training going on, the performances are progressively improved and stabilized within a relatively large range; 3) the training with more hard-class samples, i.e. $\alpha>0$ and $\beta>1$ could effectively improve the performances.

\section{Conclusion and Future Works}
In this work, we systematically investigated the hard class problem in ZSL. Firstly, we empirically found that some unseen classes disproportionally affect the classification performance in ZSL than others, and it is seemingly independent of the used ZSL methods. Then, we investigated the characteristics and possible causes of hard classes, as well as the potential benefit by taking the hard class information into account during training. Based on such analyses, we proposed some metrics to identify hard classes in both the inductive and transductive settings. With the identified hard classes, we proposed two frameworks to further boost ZSL performance, namely HarS for inductive ZSL and HarST for transductive ZSL. The proposed frameworks could accommodate most existing ZSL methods to improve their performances with marginal efforts. In this paper, we instantiated each framework with two typical ZSL methods and validated the proposed methods on three popular benchmarks. Extensive results showed that the proposed frameworks can effectively boost existing ZSL methods and the instantiated methods achieved new state-of-the-art performances. In addition, a comprehensive analysis was provided to assess the effectiveness of the key components in the proposed frameworks.

Finally we would point out that the hard class problem raised in this paper is still an under-explored problem, further efforts are needed in future. The following two issues are of particular importance of further in-depth investigation: 1) how to design appropriate metrics for identifying hard classes; 2) how to inject hard-class information into the ZSL training. For the first one, for example, the hard-class metric could be designed according to not only semantic information but also visual information in the inductive setting or by combining static semantic similarity and dynamic model predictions in the transductive setting. For the second one, how to design suitable loss functions is particularly worthy of pursuing, which could highlight hard classes for both visual feature learning based methods or generative methods. Besides, it is also expected that the quality of pseudo labels can be further improved by using two or multiple complementary base ZSL models in the transductive framework. In sum, in this work, we highlighted the importance of the hard class problem and its possible causes. However, how to improve the ZSL performance by fully exploiting hard classes is still largely an open topic, and our currently proposed remedies are merely some tentative attempts toward this goal.


%



\ifCLASSOPTIONcompsoc
  \section*{Acknowledgments}
\else
  \section*{Acknowledgment}
\fi

This work was supported by the National Natural Science Foundation of China (NSFC) under Grants (61991423, U1805264) and the Strategic Priority Research Program of the Chinese Academy of Sciences (XDB32050100).

\ifCLASSOPTIONcaptionsoff
  \newpage
\fi

\bibliographystyle{IEEEtran}
\bibliography{ref}
\begin{IEEEbiography}[{\includegraphics[width=1in,height=1.25in,clip,keepaspectratio]{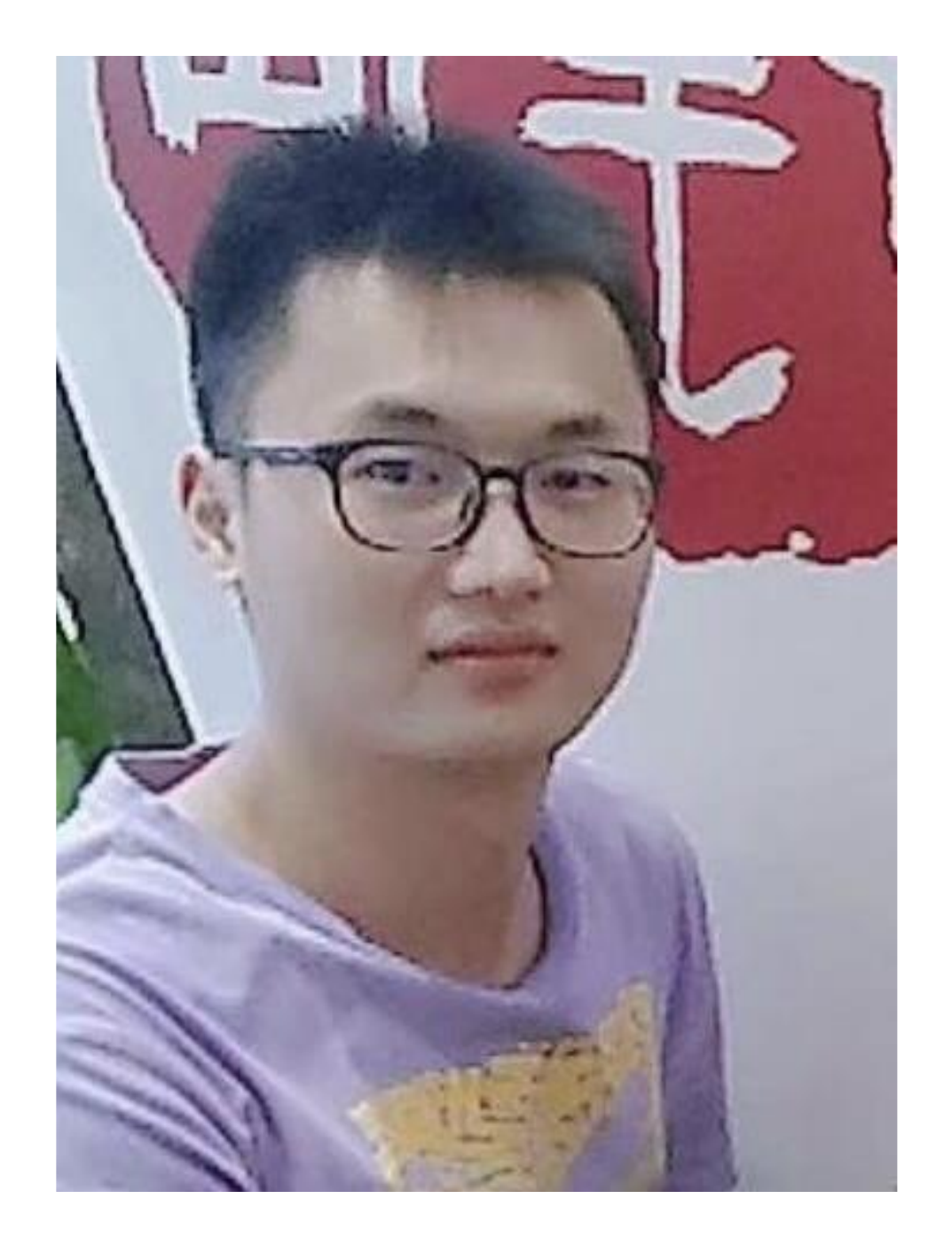}}]{Bo Liu} is currently a Ph.D. student in Pattern Recognition and Intelligence Systems at the National Laboratory of Pattern Recognition, Institute of Automation, Chinese Academy of Sciences. His research interests are in machine learning, in particular zero-shot learning and its applications in image recognition and 3D scene understanding.
\end{IEEEbiography}

\begin{IEEEbiography}[{\includegraphics[width=1in,height=1.25in,clip,keepaspectratio]{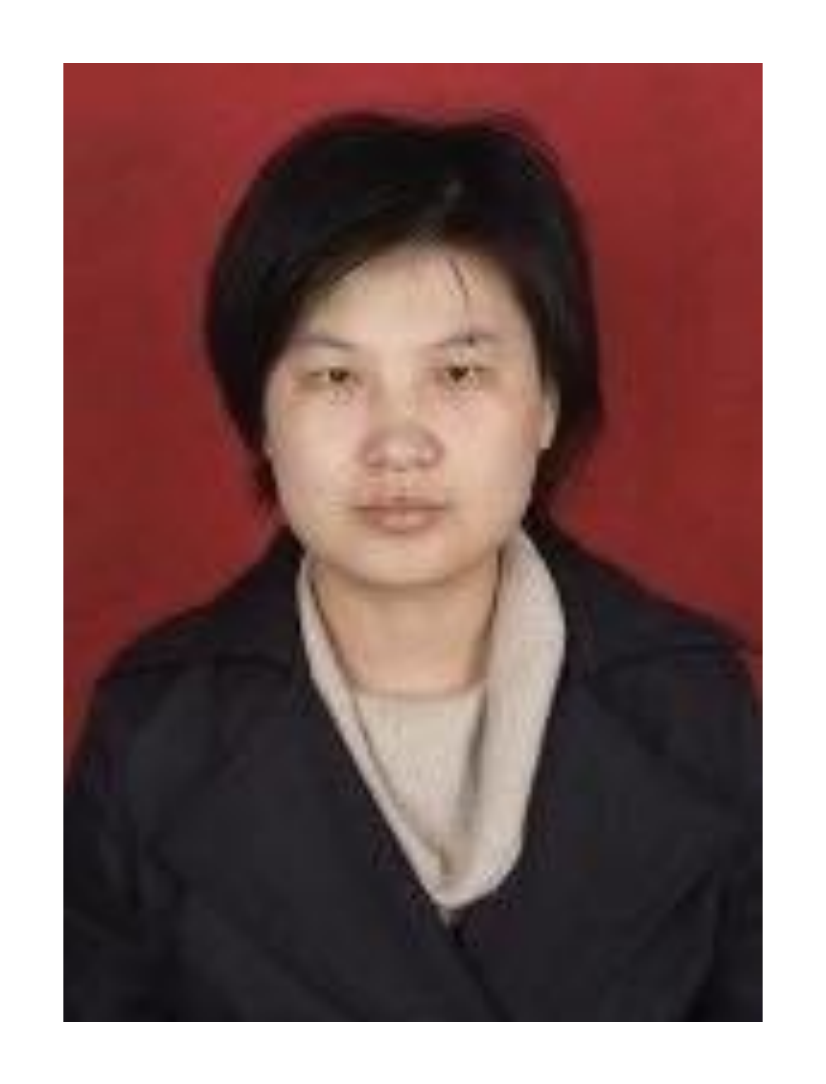}}]{Lihua Hu} received the B.S., M.S., and Ph.D. degrees all from Taiyuan University of Science and Technology. She is now an associate professor in Taiyuan University of Science and Technology. Her main research interests include visual metrology, image-based pose estimation and 3D scene understanding.
\end{IEEEbiography}

\begin{IEEEbiography}[{\includegraphics[width=1in,height=1.25in,clip,keepaspectratio]{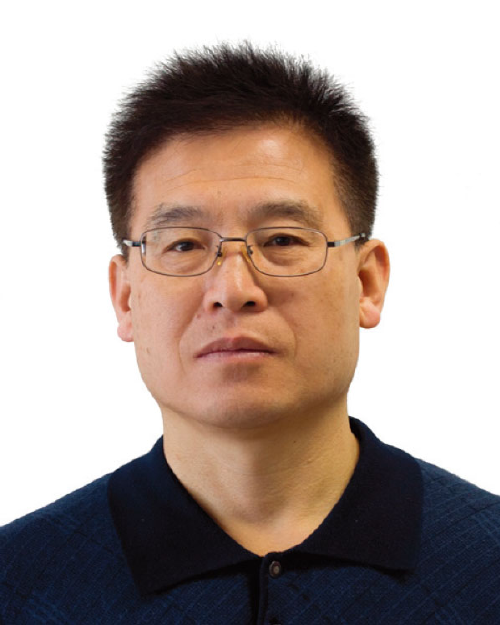}}]{Zhanyi Hu} received his B.S. in Automation from the North China University of Technology, Beijing, China, in 1985, and the Ph.D. in Computer Vision from the University of Liege, Belgium, in 1993. Since 1993, he has been with the National Laboratory of Pattern Recognition at Institute of Automation, Chinese Academy of Sciences, where he is now a professor. His research interests are in robot vision. He was a local co-chair of ICCV 2005, a PC co-chair of ACCV 2012.
\end{IEEEbiography}

\begin{IEEEbiography}[{\includegraphics[width=1in,height=1.25in,clip,keepaspectratio]{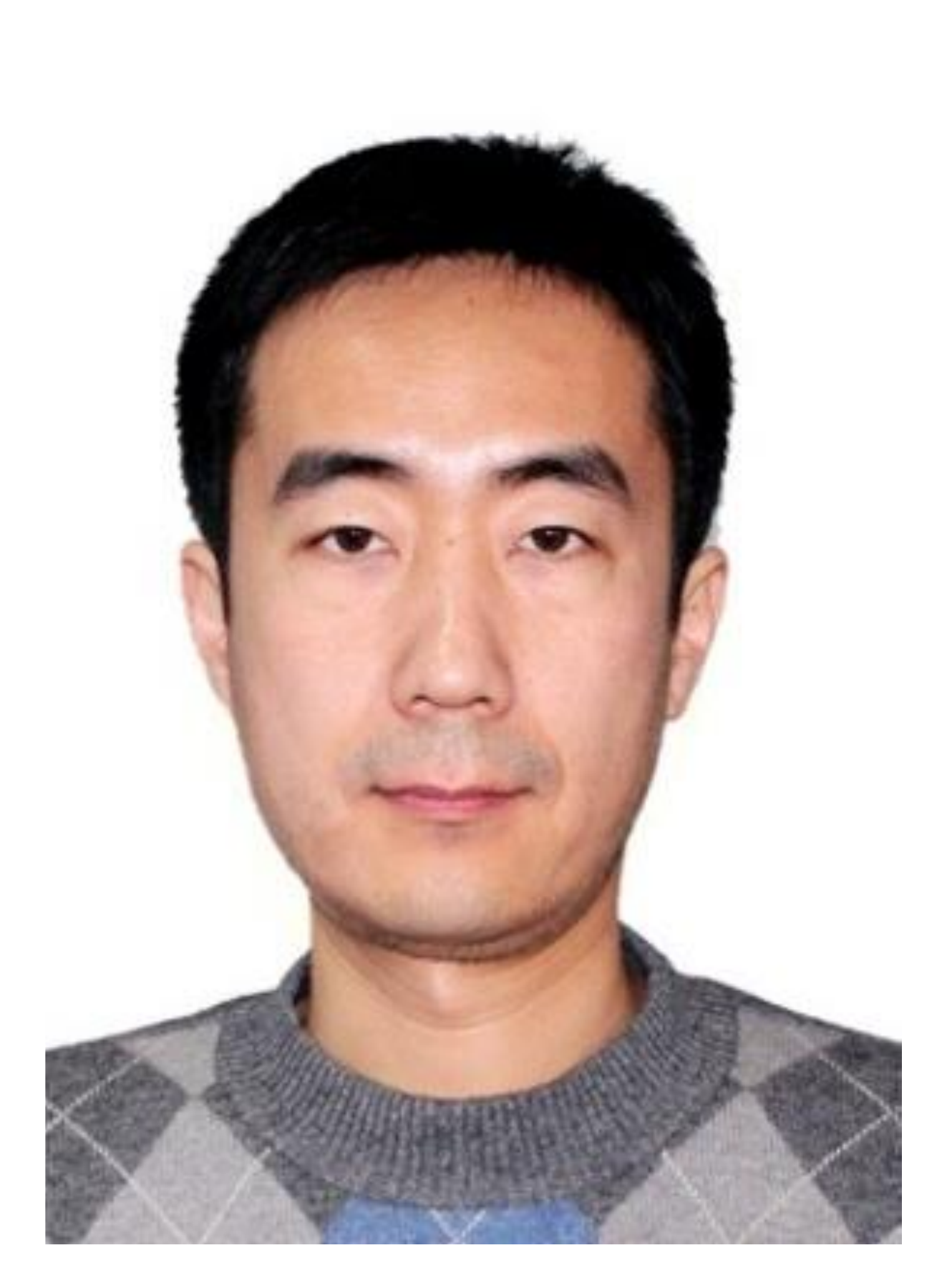}}]{Qiulei Dong} is currently a professor in the National Laboratory of Pattern Recognition, Institute of Automation, Chinese Academy of Sciences, an adjunct professor in the School of Artificial Intelligence, University of Chinese Academy of Sciences, and an adjunct professor at the Center for Excellence in Brain Science and Intelligence Technology, Chinese Academy of Sciences. He also serves as a young associate editor of the Journal of Computer Science and Technology. His research interests include 3D computer vision, pattern classification.
\end{IEEEbiography}

\end{document}